\documentclass[final,5p,times,twocolumn]{elsarticle}

\makeatletter
\renewcommand\section{\@startsection {section}{1}{\z@}%
                                   {-4ex \@plus -1ex \@minus -.2ex}%
                                   {0.5ex \@plus .2ex}%
                                   {\normalfont\large\bfseries}}
\renewcommand\subsection{\@startsection{subsection}{2}{\z@}%
                                     {-3.75ex\@plus -1ex \@minus -.2ex}%
                                     {0.5ex \@plus .2ex}%
                                     {\normalfont\normalsize\bfseries}}
\makeatother

\usepackage{float} 

\usepackage{amssymb}
\usepackage{newtxtext,newtxmath} 
\usepackage{graphicx}
\usepackage[export]{adjustbox}
\usepackage{url}
\usepackage{latexsym}
\usepackage{algpseudocode}
\usepackage{amsmath}
\usepackage{lipsum}
\usepackage{caption}
\usepackage{bbding}
\usepackage{wasysym}
\usepackage{blindtext}
\usepackage{booktabs}
\usepackage{tabularx}
\usepackage{array}
\usepackage{pifont}
\usepackage[export]{adjustbox}
\usepackage{float}

\usepackage{tabularx}
\usepackage{multirow}
\usepackage{array}

\usepackage{xcolor}

\makeatletter
\def\ps@pprintTitle{%
    \let\@oddhead\@empty
    \let\@evenhead\@empty
    \def\@oddfoot{\hbox to \textwidth{\hfil\thepage\hfil}}%
    \let\@evenfoot\@oddfoot
}
\makeatother

\begin{document}

\begin{frontmatter}

\title{Abstractive Text Summarization: State of the Art, Challenges, and Improvements}


\author[inst1]{Hassan Shakil\corref{cor1}}
\ead{hshakil@uccs.edu}
\author[inst2]{Ahmad Farooq}
\ead{afarooq@ualr.edu}
\author[inst1]{Jugal Kalita}
\ead{jkalita@uccs.edu}

\cortext[cor1]{Corresponding Author: Hassan Shakil}
 
\affiliation[inst1]{organization={Department of Computer Science,  University of Colorado },
            city={Colorado Springs},
            postcode={80918}, 
            state={CO},
            country={USA}}

\affiliation[inst2]{organization={Department of Electrical and Computer Engineering, University of Arkansas},
            city={Little Rock},
            postcode={72204}, 
            state={AR},
            country={USA}}



\begin{abstract}

Specifically focusing on the landscape of abstractive text summarization, as opposed to extractive techniques, this survey presents a comprehensive overview, delving into state-of-the-art techniques, prevailing challenges, and prospective research directions. We categorize the techniques into traditional sequence-to-sequence models, pre-trained large language models, reinforcement learning, hierarchical methods, and multi-modal summarization. Unlike prior works that did not examine complexities, scalability and comparisons of techniques in detail, this review takes a comprehensive approach encompassing state-of-the-art methods, challenges, solutions, comparisons, limitations and charts out future improvements - providing researchers an extensive overview to advance abstractive summarization research. We provide vital comparison tables across techniques categorized - offering insights into model complexity, scalability and appropriate applications.
The paper highlights challenges such as inadequate meaning representation, factual consistency, controllable text summarization, cross-lingual summarization, and evaluation metrics, among others. Solutions leveraging knowledge incorporation and other innovative strategies are proposed to address these challenges. The paper concludes by highlighting emerging research areas like factual inconsistency, domain-specific, cross-lingual, multilingual, and long-document summarization, as well as handling noisy data. Our objective is to provide researchers and practitioners with a structured overview of the domain, enabling them to better understand the current landscape and identify potential areas for further research and improvement.
\end{abstract}





\end{frontmatter}


\section{Introduction}
\label{}
The need for automatic summarization has increased substantially with the exponential growth of textual data. Automatic summarization generates a concise document that contains key concepts and relevant information from the original document \cite{maybury1999advances, nallapati-etal-2016-abstractive}. Based on the texts of the generated summaries, we can characterize summarization into two types: extractive and abstractive. In extractive text summarization, the generated summary is made up of content directly extracted from the source text \cite{widyassari2020review}, whereas in abstractive text summarization, the concise summary contains the source text's salient ideas in the newly generated text. The generated summary potentially contains different phrases and sentences that are not present in the original text \cite{nallapati-etal-2016-abstractive}. 

Although the extractive method has long been used for summary generation, the abstractive approach has recently gained popularity because of its ability to generate new sentences that better capture the main concepts of the original text, mimicking how humans write summaries. This change in emphasis is due to the maturity of extractive summarization techniques and the desire to push boundaries and address capability limitations, which leaves the dynamic and largely uncharted field of abstractive summarization open to further research and advancement \cite{widyassari2020review}.

To set the stage for continued progress in this emerging field, it is crucial to outline the characteristics that make an automatic summary not just functional but exceptional. A high-quality automatically generated summary should possess the following properties \cite{el2021automatic, ferreira2014multi, kryscinski-etal-2019-neural, jevzek2008automatic}:

\begin{itemize}

\item Concise: A high-quality summary should effectively convey the most important information from the original source while keeping the length brief.

\item Relevant: The information presented in the summary should be relevant to the main topic.

\item Coherent: A good summary should have a clear structure and flow of ideas that make it easy to understand and follow.

\item Accurate: The summary's information should be factually correct and should not contain false or misleading information.

\item Non-redundant: Summary sentences should not repeat information.

\item Readable: The sentences used in the summary should be easily understood by the target audience.

\item Fair: The summary should present the information objectively and without bias, maintaining an impartial perspective and avoiding any tonal or ideological leanings.


\item Consistent: The summary should be consistent with the original source in terms of style, tone, and format.

\item Resilience to input noise: The summary should be accurate and coherent despite noisy or poorly structured input text.

\item Multilingual capability: The summary should be able to be generated in various languages to meet the demands of a worldwide audience.

\item Adaptability to different output formats: The summary should be adaptable enough to be output in a variety of formats, including bullet points, paragraph summaries, and even visual infographics.

\end{itemize}
A generated abstractive summary provides an accurate, concise, and easy-to-understand representation of the source text by fulfilling these properties and doing so in its own words.

Abstractive summarization encounters various challenges in its quest to generate new text for a condensed and cohesive summary of the original text. The resulting summary maintains the same style and tone as the original text while ensuring factual and grammatical accuracy, fluency, and coherence. In addition, it can be difficult to summarize text that contains varying opinions and perspectives while still being concise and informative. This survey provides a comprehensive overview of the current state of the art in abstractive summarization, including the latest techniques, recent improvements that have been accomplished, issues, and challenges that need to be addressed. We categorize the state-of-the-art techniques into five groups based on the underlying methodologies and techniques used in text summarization approaches. Each category represents a distinct group of methods that share common features or principles. Traditional Sequence-to-Sequence (Seq2Seq) models \cite{sutskever2014sequence, shi2021neural} leverage encoder-decoder architectures to map input texts to summarized outputs. A fruitful approach involves using pre-trained Large Language Models (pre-trained LLMs) \cite{devlin2018bert, syed2022survey}, which capitalize on large-scale (unsupervised) training to capture general contextual and linguistic information, and then further specialized training to generate effective summaries. Reinforcement Learning (RL) approaches \cite{sutton2018reinforcement, ryang2012framework} also play a significant role, with models learning to optimize summary quality based on human-like preferences. Hierarchical approaches \cite{yang2016hierarchical, fabbri2019multi}, on the other hand, focus on exploiting the inherent structure of input texts to generate more coherent and informative summaries. Finally, Multi-modal Summarization \cite{jangra2023survey, zhang2022unims} methods combine different data modalities, such as text and images, to generate comprehensive and context-rich summaries. By categorizing the state-of-the-art techniques in this way, we hope to provide a clear and structured overview of the current research landscape in abstractive summarization.

The integration of structured knowledge bases has significantly improved the accuracy and coherence of content generated by natural language models \cite{ratnaparkhi-2000-trainable, bordes2011learning}. Despite this, challenges in capturing intricate meanings remain, leading to innovations like attention mechanisms \cite{bahdanau2014neural} and advanced knowledge representation \cite{lin2018knowledge}. Traditional metrics often miss the semantic depth and have encouraged the adoption of BERTScore \cite{zhang2019bertscore} and MoverScore \cite{zhao2019moverscore}. Summarizing long documents is addressed using hierarchical models \cite{fabbri2019multi} and memory-augmented networks \cite{santoro2016meta}. Emphasis on factual consistency has spurred strategies like knowledge integration \cite{pasunuru-etal-2017-towards} and reinforcement learning \cite{zhang-etal-2020-optimizing}. The emergence of models like CTRL \cite{keskar2019ctrl} highlights the trend toward controllable summarization. With the digital realm diversifying, there is a push for multimodal summarization. As AI becomes more influential, transparency and interpretability are paramount. Overall, abstractive text summarization is continuously evolving, meeting challenges with innovative solutions.
\subsection{Prior Surveys on Abstractive Summarization}

Several prior surveys have explored the developments in automatic summarization methods. These survey papers offer vital insights into the methods, limitations, and potential future research directions in automatic summarization. A significant portion of these surveys, such as those conducted by Nazari et al. \cite{nazari2019survey}, and Moratanch et al. \cite{moratanch2017survey}, primarily focused on extractive summarization methods. This focus can be attributed to the complexity inherent in abstractive summarization.
In recent years, a growing body of work has concentrated on the state of the art in abstractive summarization. For instance, Suleiman et al. \cite{suleiman2020deep}, Zhang et al.\cite{zhang2022comprehensive}, and Gupta et al. \cite{gupta2019abstractive} have exclusively focused on abstractive text summarization. These studies delve into deep learning-based abstractive summarization methods and compare performance on widely used datasets. Lin et al. \cite{lin2019abstractive} explored existing neural approaches to abstractive summarization, while Gupta et al. \cite{gupta2019abstractive} characterized abstractive summarization strategies, highlighting the difficulties, tools, benefits, and drawbacks of various approaches. Syed et al. \cite{syed2021survey} evaluated various abstractive summarization strategies, including encoder-decoder, transformer-based, and hybrid models, and also discussed the challenges and future research prospects in the field.

There are studies that cover both extractive and abstractive methods, providing a more comprehensive view of the field. Examples of such works include Gupta et al. \cite{gupta2019text} and Mahajani et al. \cite{mahajani2019comprehensive}. These studies offer a comparative examination of the effectiveness of both extractive and abstractive techniques as well as giving an overview of both. Ermakova et al. \cite{ermakova2019survey} presented a study on the evaluation techniques utilized in summarization, which is fundamental for comprehending the viability and potential improvements in both extractive and abstractive summarization methods. These works go about as an extension between the two summarization approaches, displaying their individual benefits and possible cooperative synergies.


Previous research on automatic summarization has oftentimes focused on specific topics, for example, abstractive summarization utilizing neural approaches, deep learning-based models, sequence-to-sequence based models, and extractive summarization. Some have included the challenges and limitations of abstractive summarization, while others have focused on the best way to improve and evaluate it. 

Nonetheless, there is a need for a more comprehensive analysis that covers state-of-the-art methods, challenges, solutions, comparisons, limitations, and future improvements - providing a structured overview to advance abstractive summarization research.

Table \ref{tab:compar1} gives an overall comparison of our review with various survey papers that are available in the literature. Unlike prior studies, our work examines state-of-the-art methods, challenges, solutions, comparisons, limitations and charts out future improvements - providing researchers an extensive overview to advance abstractive summarization research. The following are the main contributions of this study:

\begin{itemize}

\item Overview of the state-of-the-art techniques in abstractive text summarization: This study incorporates information from a plethora of relevant studies to give an overview of the ongoing methodologies utilized in abstractive summarization. This can help researchers and practitioners familiarize themselves quickly with the most recent developments and patterns.

\item Comparative Analysis of Models: This study includes a comparative analysis of models in abstractive summarization, focusing on scales, training time, and resource consumption, among other categories. This unique dimension offers practical insights, aiding in the selection of efficient models and enriching the field with valuable, often overlooked information.

\item Identification of challenges in abstractive summarization: This study presents current issues and challenges in abstractive summarization by consolidating information from various research papers, for instance, the challenge of generating coherent and grammatically accurate summaries. Our work can help specialists focus on these areas and develop innovative solutions by highlighting such issues.

\item Discussion of potential improvements in abstractive summarization: This study explores strategies to enhance abstractive summarization, for example, incorporating knowledge and various techniques to generate factually accurate and coherent summaries. This can aid researchers in finding better approaches to generate high-quality abstractive summarization frameworks.

\item Exploration of future research directions: This study highlights emerging frontiers like personalized summarization, long-document summarization, multi-document summarization, multilingual capabilities, and improved evaluation metrics along with leveraging recent advances in Large Language Models (LLMs). It also highlights future directions to overcome limitations like inadequate representation of meaning, maintaining factual consistency, explainability and interpretability, ethical considerations and bias, and further related concepts to help advance the field.

\item Holistic survey of abstractive summarization: Unlike prior works focused solely on extractive summarization, this review takes a comprehensive approach, encompassing state-of-the-art abstractive summarization methods, along with comparisons and analyses of complexities, challenges, and solutions. It provides researchers with a structured overview to advance abstractive summarization research.

\end{itemize}

\newcolumntype{Y}{>{\centering\arraybackslash}X}
\newcommand{\cmark}{\ding{51}}
\newcommand{\xmark}{\ding{55}}

\renewcommand{\arraystretch}{1} 
\renewcommand\tabularxcolumn[1]{m{#1}} 

\begin{table*}
\centering
\begin{tabularx}{\textwidth}{|Y|Y|Y|Y|Y|Y|Y|}
\hline
\textbf{Paper (Ref.)} & \textbf{Year} & \textbf{Primary Focus} & \textbf{Methodologies Explored} & \textbf{Challenges Addressed} & \textbf{Improvements Suggested} & \textbf{Future Directions Highlighted} \\
\hline
Nazari et al. \cite{nazari2019survey} & 2019 & Extractive & Statistical, ML, Semantic-Based, Swarm Intelligence & \xmark & \xmark & \cmark \\
\hline
Moratanch et al. \cite{moratanch2017survey} & 2017 & Extractive & Supervised, Unsupervised & \cmark & \xmark & \cmark \\
\hline
Gupta et al. \cite{gupta2019abstractive} & 2019 & Both & Conventional Algorithmic Approaches, Domain-Specific & \cmark & \xmark & \cmark \\
\hline
Mahajani et al. \cite{mahajani2019comprehensive} & 2019 & Both & Traditional Seq2Seq, RL, Hierarchical & \xmark & \xmark & \cmark \\
\hline
Suleiman et al. \cite{suleiman2020deep} & 2020 & Abstractive & Traditional Seq2Seq, Pre-trained Large Language Models, RL & \cmark & \xmark & \xmark \\
\hline
Zhang et al. \cite{zhang2022comprehensive} & 2022 & Abstractive & Traditional Seq2Seq, Hierarchical & \cmark & \xmark & \cmark \\
\hline
Gupta et al. \cite{gupta2019abstractive} & 2019 & Abstractive & Structure-based, Semantic-based, Deep Learning and neural network based & \cmark & \xmark & \cmark \\
\hline
Lin et al. \cite{lin2019abstractive} & 2019 & Abstractive & Traditional Seq2Seq, Hierarchical, RL & \xmark & \xmark & \xmark \\
\hline
Syed et al. \cite{syed2021survey} & 2021 & Abstractive & Traditional Seq2Seq, Pre-trained Large Language Models, Hierarchical & \cmark & \xmark & \cmark \\
\hline
Ermakova et al. \cite{ermakova2019survey} & 2019 & Evaluation Methods & - & \cmark & \xmark & \xmark \\
\hline
Our Survey & 2024 & Abstractive & Traditional Seq2Seq, Pre-trained Language Models, RL, Hierarchical, Multi-modal & \cmark & \cmark & \cmark \\
\hline
\end{tabularx}
\caption{Comparison of Our Survey with Existing Survey Articles}
\label{tab:compar1}
\end{table*}

\subsection{Organization}
In this paper, we present a comprehensive survey of abstractive summarization, encompassing the state of the art, challenges, and advancements. Section II delves into automatic summarization, detailing its various types along with examples. Section III reviews the literature on the state of the art in abstractive summarization. Section IV explores model scalability and computational complexity in abstractive summarization. Section V addresses issues, challenges, and future directions for abstractive summarization. The concluding remarks are offered in Section VI.

\section{Automatic Summmarization}
Automatic summarization is a technique used in Natural Language Processing (NLP) to generate a condensed version of a longer text document while retaining the most important information \cite{el2021automatic}. The aim of automatic summarization is to reduce the length of the text without losing the essence of the source content. The primary purpose of summarization is to help people get a quick understanding of the main topics and ideas covered in a large text document without having to read the entire document. As mentioned at the beginning of the paper, there are two main types of automatic summarization: extractive and abstractive summarization \cite{nazari2019survey}. A fundamental comparison between extractive and abstractive summarization techniques is presented in Table \ref{tab:extractive_abstractive_comparison}.

\begin{table*}  
\centering
\renewcommand\tabularxcolumn[1]{m{#1}} 
\begin{tabularx}{\textwidth}{|Y|Y|Y|}  
\hline
\textbf{Criteria} & \textbf{Extractive Summarization} & \textbf{Abstractive Summarization} \\ 
\hline
Overview & Selecting key sentences or phrases from the original text & Generating a summary that includes new sentences that capture the essence of the source text \\ 
\hline
Output Length & Fixed (number of sentences or words) & Variable (can be adjusted to meet specific requirements) \\ 
\hline
Readability & Can be choppy and lack coherence & Can generate a more fluent and coherent summary \\ 
\hline
Information & Preserves important information from the original text & Can introduce new sentences and reorganize information from the original text \\ 
\hline
Performance & Generally faster and less resource-intensive & More complex and computationally expensive \\ 
\hline
Evaluation & ROUGE, precision/recall, F1-score & ROUGE, human evaluation, coherence and fluency measures \\ 
\hline
Examples & TextRank\cite{mihalcea-tarau-2004-textrank}, LexRank\cite{erkan2004lexrank}, Latent Semantic Analysis\cite{gong2001generic} & Pointer-Generator Network\cite{46111}, Transformer-based models\cite{vaswani2017attention}, BERT-based models\cite{devlin-etal-2019-bert} \\ 
\hline
\end{tabularx}
\caption{Comparison of Extractive and Abstractive Summmarization}
\label{tab:extractive_abstractive_comparison}
\end{table*}

\subsection{Extractive Summarization} 
Extractive summarization is the process that entails cherry-picking the most salient sentences or phrases from the source text and fusing them into a summary \cite{liang2021improving}. The chosen sentences or phrases typically include important details pertaining to the subject being discussed. Extractive summarization refrains from any form of paraphrasing or rewriting of the source text. Instead, it highlights and consolidates the text's most crucial information in a literal way. For instance, Google News utilizes an extractive summarization tool that sifts through news articles and generates a summary by pulling out the most relevant sentences \cite{bhatia2015literature}. Table \ref{cap1} presents an extractive summary of a source text, which was obtained from the United Nations Climate Action website\footnote{https://www.un.org/en/climatechange/cop26}. This summary was generated by the ChatGPT-4 model, a product of OpenAI\footnote{https://openai.com/gpt-4}.

\begin{table} [ht]
    \centering
    \setlength{\tabcolsep}{6pt} 
    \renewcommand{\arraystretch}{1.5} 
    \begin{tabular}{|m{0.45\textwidth}|m{0.45\textwidth}|} 
        \hline
        \textbf{Source Text:} Following weeks of tense negotiations, the international climate conference in New Delhi, India, finally concluded yesterday. The conference was attended by delegates from 197 countries, all of whom pledged to reduce carbon emissions drastically in the next decade. The resulting agreement, referred to as the `New Delhi Accord', mandates that signatory countries cut their emissions by 50\% by the year 2033. This target represents a significantly larger commitment than previous agreements. The accord also includes provisions for financial assistance to developing countries, helping them transition to greener technologies. Critics, however, argue that the targets are unrealistic and warn of the potential economic repercussions.\\
        \hline
        \textbf{Extractive Summary:} Following weeks of tense negotiations, the international climate conference in New Delhi, India, finally concluded yesterday. The conference was attended by delegates from 197 countries. The resulting agreement, referred to as the `New Delhi Accord', mandates that signatory countries cut their emissions by 50\% by the year 2033. The accord also includes provisions for financial assistance to developing countries. Critics, however, argue that the targets are unrealistic and warn of the potential economic repercussions.\\
        \hline
        \textbf{Abstractive Summary:} After lengthy discussions, the global climate meeting in India produced a landmark agreement known as the `New Delhi Accord'. With the involvement of nearly 200 countries, the accord sets ambitious goals for halving carbon emissions by 2033 and supporting developing nations financially to embrace environmentally friendly technologies. Despite the bold steps, some observers express concerns over the feasibility of these goals and their potential economic impact.\\
         \hline
    \end{tabular}
\caption{Example of Extractive and Abstractive Summarization}
\label{cap1}
\end{table}

\subsection{Abstractive Summarization}
Abstractive summarization involves creating a summary that is not just a selection of sentences or phrases from the source text, but is compromised of newly minted sentences that capture the essence of the original text \cite{kryscinski-etal-2020-evaluating}. The model generates new sentences that maintain the original text's meaning but are usually shorter and more to the point in order to achieve the abstraction of ideas. Abstractive summarization is more complex than extractive summarization because it necessitates the NLP model to comprehend the text's meaning and generate new sentences. The New York Times summary generator, which generates summaries that are very similar to those written by humans, is a great example of abstractive summarization. Table \ref{cap1} showcases an abstractive summary of the source text, sourced from the United Nations Climate Action website. This summary was synthesized using the ChatGPT-4 model, mentioned earlier.

\section{State of the Art in Abstractive Text Summarization}

\begin{figure*}[htbp]
    \centering
    \includegraphics[width=\linewidth, frame]{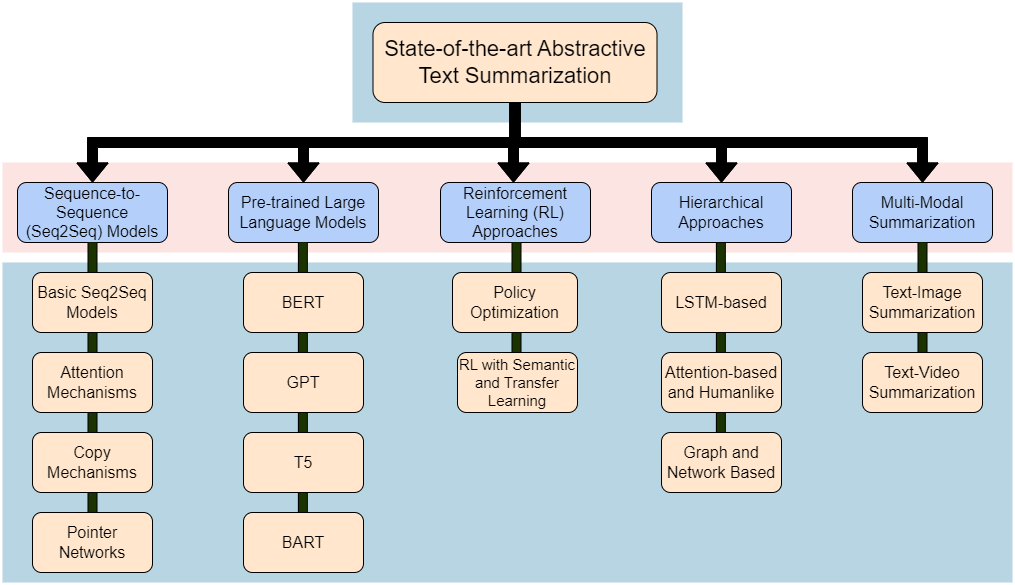}
    \caption{Taxonomy of State-of-the-art Abstractive Text Summarization}
    \label{fig:TaxonomyATSS}
\end{figure*}

We present a comprehensive taxonomy of state-of-the-art abstractive text summarization based on the underlying methods and structures found in the literature; see Figure \ref{fig:TaxonomyATSS}. At a fundamental conceptual level, summarization consists of transforming a long sequence of sentences or paragraphs into a concise sequence of sentences. Thus, all machine learning models that learn to perform summarization can be characterized as Sequence-to-Sequence (Seq2Seq) models. However, Seq2Seq models encompass a wide variety of approaches, one of which we call Traditional Seq2Seq models in the taxonomy. We classify state-of-the-art abstractive text summarization into five distinct categories: Traditional Sequence-to-Sequence (Seq2Seq) based Models \cite{sutskever2014sequence}, Pre-trained Large Language Models \cite{devlin2018bert, syed2022survey}, Reinforcement Learning (RL) Approaches \cite{sutton2018reinforcement, ryang2012framework}, Hierarchical Approaches \cite{yang2016hierarchical, fabbri2019multi}, and Multi-modal Summarization \cite{jangra2023survey, zhang2022unims}. Although distinguishing between approaches and systems in abstractive text summarization can be difficult, our taxonomy strives to provide a clear and well-defined division. In addition, most state of the art methods possess subclasses, as depicted in Figure
\ref{fig:TaxonomyATSS}. This organized perspective on state-of-the-art abstractive text summarization methods enables researchers and practitioners to better understand the current landscape of the field and identify potential areas for further research and improvement.

\subsection{Traditional Sequence-to-Sequence (Seq2Seq) Models}

\begin{figure*}[htbp]
    \centering
    \includegraphics[width=\linewidth, frame]{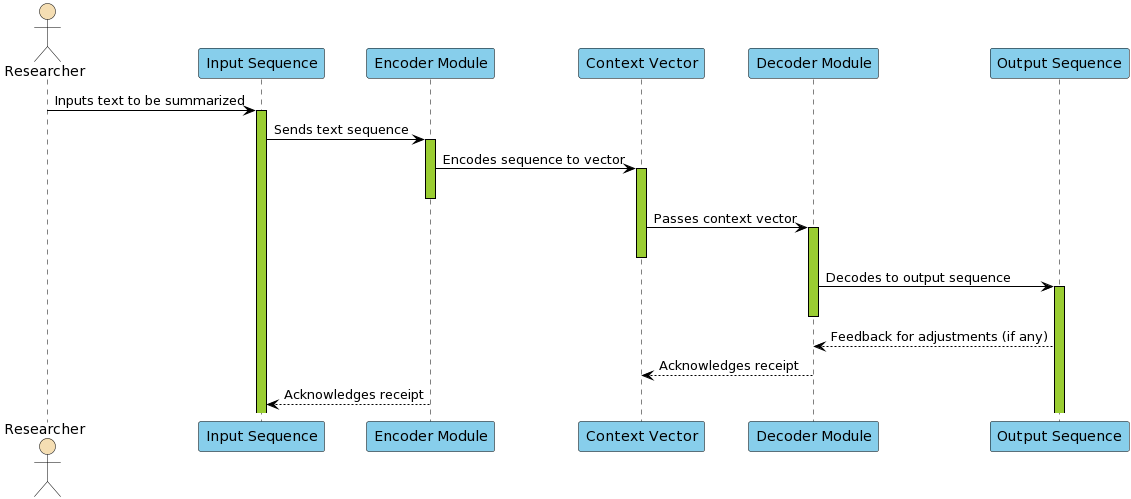}
    \caption{Traditional Seq2Seq model flow for abstractive text summarization}
    \label{fig: Seq2SeqDia}
\end{figure*}

Traditional Seq2Seq models are a class of neural network architectures developed to map input sequences to output sequences. They are frequently employed in tasks involving the processing of natural languages, such as summarization and machine translation. The fundamental concept is to first use an encoder to turn the input sequence into a fixed-length vector representation and then to use a decoder to generate the output sequence from the vector representation \cite{sutskever2014sequence}. The sequence diagram in Figure \ref{fig: Seq2SeqDia} illustrates the flow of traditional Seq2Seq models, emphasizing their significance and applications in the realm of abstractive text summarization. To provide a more comprehensive understanding of Seq2Seq models in the context of abstractive text summarization, we have further divided them into four sub-classes: Basic Seq2Seq Models, Attention Mechanisms, Copy Mechanisms, and Pointer Networks. This classification makes it possible to give a clearer analysis of the various methods used in Seq2Seq models for abstractive text summarization. Table \ref{tab:seq2seq_variations} shows a comparison of various sub-classes of Traditional Sequence-to-Sequence (Seq2Seq) Models..

\begin{table*}
\centering
\begin{tabularx}{\textwidth}{|Y|Y|Y|Y|}
\hline
\textbf{Sub-class} & \textbf{Architecture \& Mechanism} & \textbf{Strengths \& Contributions} & \textbf{Weaknesses} \\
\hline
Basic Seq2Seq & Multilayered LSTM for sequence mapping & Efficient mapping, good with long sentences, introduced end-to-end sequence learning & Struggles with out-of-vocabulary words \\
\hline
Attention Mechanisms & Attention-based model focusing on relevant input parts & Improved focus and scalability, data-driven abstractive summarization approach & May miss complex dependencies \\
\hline
Copy Mechanism & COPYNET integrating word generation with copying & Effective replication, excels in summarization, demonstrated copying efficacy & Requires mechanisms for copying decisions \\
\hline
Pointer Networks & Ptr-Net with attention as a pointer for variable output size & Solves complex problems, excellent generalization, introduced solution for variable output size & Best for discrete token output tasks \\
\hline
\end{tabularx}
\caption{Comparison of sub-classes of Traditional Sequence-to-Sequence (Seq2Seq) Models}
\label{tab:seq2seq_variations}
\end{table*}

\subsubsection{Basic Seq2Seq models}
Basic Seq2Seq models, first introduced by Sutskever et al. \cite{sutskever2014sequence}, utilize the encoder-decoder architecture to generate summaries. Although machine translation was the model's primary application, the Seq2Seq framework has since been used for abstractive text summarization. A convolutional neural network (CNN)-based Seq2Seq model for natural language phrases was proposed by Hu et al. \cite{hu2014convolutional}. Although the paper's primary focus was on matching phrases, the basic Seq2Seq model presented can be adapted for abstractive text summarization. However, these basic models face limitations in capturing long-range dependencies, leading to the development of attention mechanisms \cite{bahdanau2014neural}.

\subsubsection{Attention Mechanisms}
Attention mechanisms enable models to selectively focus on relevant parts of the input during the decoding phase, improving the generated summaries' quality.  A novel neural attention model for abstractive sentence summarization was proposed by Rush et al. \cite{rush2015neural}. The task of creating a condensed version of an input sentence while maintaining its core meaning is called a single-sentence summary, which was the focus of this study. This research is considered a trailblazer since it was among the first to use attention mechanisms for abstractive text summarization. The proposed model was based on an encoder-decoder framework. The encoder is a CNN that processes the input sentence, while the decoder is a feed-forward Neural Network Language Model (NNLM) that generates the summary. The decoder has an attention mechanism that allows it to selectively concentrate on various sections of the input text while generating the summary. As a result, the model may learn which words or phrases of the input text are crucial for generating the summary. A ``hard" attention model and a ``soft" attention model were the two variations of the attention model tested by the authors. The soft attention model computes a weighted average of the input words, with the weights representing the relevance of each word to the summary, whereas the hard attention model stochastically chooses a restricted group of input words to be included in the summary. The soft attention model performed better in the experiments because it makes it easier for gradients to flow during training. The proposed model was evaluated on the Gigaword dataset \cite{napoles2012annotated}, a large-scale corpus of news articles with associated headlines. The results showed that the attention-based model outperformed a number of baselines, including the fundamental Seq2Seq model and a state-of-the-art extractive summarization system. The experiments also showed that the model could generate concise and coherent summaries that capture the core idea of the input text. 

An attentive encoder-decoder architecture for abstractive sentence summarization was proposed by Chopra et al. \cite{chopra2016abstractive}, focusing on generating abstractive summaries for single sentences. By using Recurrent Neural Networks (RNNs) as the foundation for both the encoder and decoder components in the Seq2Seq model, the paper extended the earlier work by Rush et al. \cite{rush2015neural}. The bidirectional RNN encoder in the attentive encoder-decoder architecture analyzes the input sentence, and the RNN decoder with an attention mechanism generates the summary. The forward and backward contexts of the input sentence are both captured by the bidirectional RNN encoder, leading to a more thorough grasp of the sentence structure. While generating each word in the summary, the model may dynamically focus on various portions of the input phrase because of the attention mechanism in the decoder. This selective focus allows the model to generate coherent and meaningful summaries. The Gigawaord dataset was used to evaluate the performance of the proposed model in comparison to various baselines, such as the basic Seq2Seq model \cite{sutskever2014sequence} and the attention-based model \cite{rush2015neural}. The results demonstrated that the attentive RNN-based encoder-decoder design generated more accurate and informative abstractive summaries and outperformed the baselines.

Nallapati et al. \cite{nallapati2016abstractive} investigated various techniques to improve the basic Seq2Seq model to advance abstractive text summarization. The authors addressed multiple aspects of the model, such as the encoder, attention mechanisms, and the decoder, to improve the model's overall performance in generating abstractive summaries. They proposed several modifications to the encoder including the incorporation of a bidirectional RNN encoder to capture both forward and backward contexts in the input text. They also used a hybrid word-character encoder to handle out-of-vocabulary (OOV) words and improve generalization. The authors explored both local and global strategies for the attention mechanism that allowed the model to concentrate on relevant input while generating the summary. Additionally, the paper introduced a switch mechanism for the decoder that enabled the model to choose between generating words based on the context vector and directly copying words from the input text. This method strengthens the model's capacity to generate coherent summaries and help tackle the issues of OOV words. On the Gigaword and DUC-2004 \cite{over2004introduction} datasets, the authors evaluated the performance of the model in comparison to a number of benchmark models. The results demonstrated that the suggested modifications improved the functionality of the basic Seq2Seq model, leading to more precise and insightful abstractive summaries.

A graph-based attention mechanism for abstractive document summarization was introduced by Tan et al. \cite{tan2017abstractive} that takes into account the relationships between sentences in a document. Traditional Seq2Seq models with attention mechanisms frequently concentrate on the words within a sentence but are unable to recognize the inter-sentence dependencies. The authors aimed to overcome this restriction by embedding the structural information of the document into the attention mechanism. The suggested approach first creates a sentence graph that represents the document, where the nodes are the sentences and the edges are the relationships between them. The attention mechanism then works on this graph, permitting the model to focus on both local and global sentence-level information. The model consists of a bidirectional RNN encoder to capture a representation of the input sentence and a decoder with a graph-based attention mechanism for generating the summary. On the CNN/Daily Mail and DUC-2004 datasets, the authors assessed the performance of the model in comparison with various state-of-the-art abstractive and extractive summarization models. The results showed that by precisely capturing the connections between sentences in a document, the graph-based attention mechanism improved the quality of generated summaries.


\subsubsection{Copy Mechanism}
Gu et al. \cite{gu2016incorporating} introduced copy mechanism, a novel approach for abstractive text summarization that addresses the challenge of handling rare and OOV words. OOV words, which are often included in real-world text data, make it difficult for conventional Seq2Seq models to generate accurate summaries. The authors provided a technique to address this problem that enables the neural network to selectively copy words from the input text straight into the generated summary. The copying mechanism was incorporated into the existing Seq2Seq framework, specifically into the attention mechanism. By removing the difficulties brought on by rare and OOV words, this method improves the model's capacity to generate more precise and coherent summaries. According to the experimental findings, adding a copy mechanism to Seq2Seq models considerably enhances their ability to perform these tasks when compared to more conventional Seq2Seq models without a copy mechanism.

For abstractive text summarization, Song et al. \cite{song2018structure} suggested a unique structure-infused copy mechanism that uses both syntactic and semantic information from the input text to help the copying process. The primary motivation behind this approach is to improve the coherence and accuracy of the generated summaries by leveraging the structural information inherent in the input text. The structure-infused copy mechanism incorporates semantic information such as named entities and key phrases along with a graph-based representation of the input text's syntactic structure, notably the dependency parse tree. The model can more accurately detect salient information and thus generate summaries that more accurately capture the major concepts of the original text by including these structural features in the copying mechanism. The authors employed a multi-task learning framework that jointly learns to generate abstractive summaries and forecast the structural characteristics of the original text. This method enables the model to capture the interdependencies between the goal of generating summaries and the task of structure prediction, generating summaries that are more accurate and coherent. On the CNN/Daily Mail dataset, the authors evaluated their structure-infused copy mechanism and contrasted its effectiveness with various state-of-the-art summarization models. The experimental findings showed that, in terms of ROUGE  \cite{lin2004rouge} scores, their strategy outperforms the baseline models.

\subsubsection{Pointer Networks}
Vinyals et al. \cite{vinyals2015pointer} proposed an enhancement to the Seq2Seq models by incorporating attention and pointers. This Pointer-Generator Network is useful for summarization tasks because it can either generate words from a preset vocabulary or copy them directly from the source, effectively handling rare or OOV words and leading to better abstractive summarization. Additionally, a coverage mechanism is integrated to monitor attention history, ensuring diverse attention and reducing redundancy in the summaries. The model's effectiveness was evaluated on the CNN/Daily Mail dataset with notable improvements in ROUGE scores.

SummaRuNNer, introduced by Nallapati et al. \cite{nallapati2017summarunner}, uses Pointer Networks for abstractive tasks in addition to being designed for extractive summarization. It extracts the top-ranked sentences for the summary by evaluating the input documents at the word and sentence levels using a hierarchical RNN. By allowing direct copying from the source text in its abstractive variant, the pointer mechanism overcomes the limitations of conventional sequence-to-sequence models. The pointer mechanism and RNN-based hierarchy work together to improve the model's summarization performance. The models were evaluated on the DUC 2002\footnote{https://www-nlpir.nist.gov/projects/duc/guidelines/2002.html} dataset by using various variants of the Rouge metric and contrasted with cutting-edge models.

Kryscinski et al. \cite{kryscinski2018improving} integrated Pointer Networks with reinforcement learning for abstractive text summarization. Their model is trained using both supervised and reinforcement learning, and it is based on the architecture of the Seq2Seq model with attention. The element of reinforcement guarantees conformity with human preferences. In order to ensure accuracy and proper handling of OOV words, the Pointer Networks incorporate straight copying from the input. Their approach performed better in terms of ROUGE scores when evaluated on the CNN/Daily Mail dataset.

Chen et al. \cite{chen2018fast} combined reinforcement learning with Pointer Networks for abstractive summarization. Their model uses reinforcement learning to optimize the process of selecting and rewriting sentences from the input. Using the CNN/Daily Mail and DUC-2002 datasets, the model was assessed using METEOR \cite{banerjee2005meteor}, standard ROUGE metrics, and human evaluations of readability and relevance.

Using a unified model, Hsu et al. \cite{hsu2018unified} presented extractive and abstractive summarization techniques. They introduced an inconsistency loss function to ensure alignment between the extractive and abstractive outputs. The abstractive component employs pointer networks for direct copying from the input, improving the quality of the summaries. Evaluated on the CNN/Daily Mail dataset, the model demonstrated strong performance through ROUGE scores and human evaluations on Amazon Mechanical Turk\footnote{https://www.mturk.com/}, assessing informativity, conciseness, and readability.

Hsu et al. \cite{hsu2018unified} introduced extractive and abstractive summarization methods using a unified model. In order to guarantee alignment between the extractive and abstractive outputs, they also implemented the inconsistency loss function. Pointer networks are used by the abstractive component to copy directly from the input, improving the summaries' quality. When the CNN/Daily Mail dataset was used for evaluation, the model performed well according to ROUGE scores and human assessments on Amazon Mechanical Turk that measured readability, and conciseness.

\subsection{Pre-trained Large Language Models}


\begin{figure*}[htbp]
    \centering
    \includegraphics[width=\linewidth, frame]{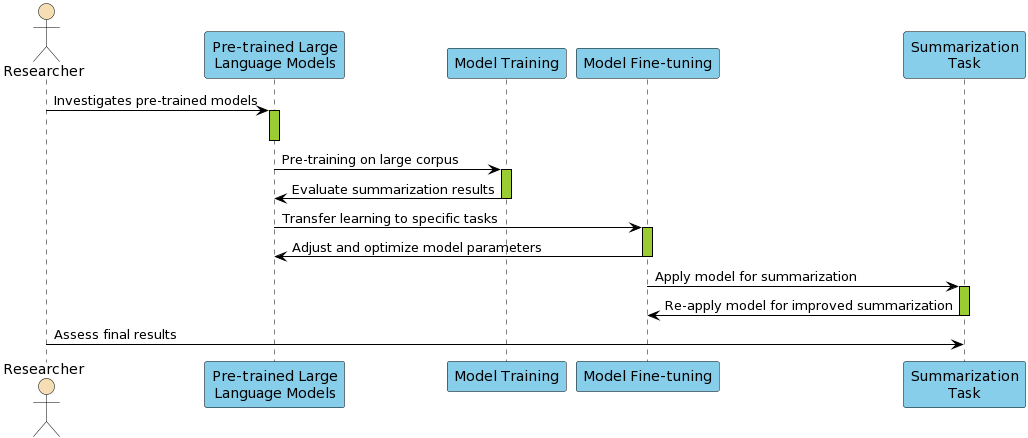}
    \caption{Pre-trained Large Language Model flow for abstractive summarization}
    \label{fig: LLM}
\end{figure*}

Pre-trained Large Language Models are large-scale neural networks that have learned contextualized representations of words, phrases, and sentences through training on enormous volumes of text data. The sequence diagram illustrated in Figure \ref{fig: LLM} shows the interaction between a researcher and a pre-trained Large Language Model (LLM), showcasing the model's ability to process queries and generate contextually relevant responses based on its extensive training. In a variety of natural language processing tasks, including abstractive text summarization, these models achieve state of the art results. To provide a more comprehensive understanding of Pre-trained Large Language models in the context of abstractive text summarization, we have further divided them into four sub-classes based on their use, as shown in Figure \ref{fig:TaxonomyATSS}: BERT (Bidirectional Encoder Representations from Transformers), GPT (Generative Pre-trained Transformer), T5 (Text-to-Text Transfer Transformer), and BART (Bidirectional and Auto-Regressive Transformers). 
The classification makes it possible to give a clearer analysis of the various methods used in Pre-trained Large Language models for abstractive text summarization. Table \ref{tab:lang_models} shows a comparison of various Pre-trained Large Language Models (the selected versions have the highest number of parameters available).

\begin{table*}
\centering
\begin{tabularx}{\textwidth}{|Y|Y|Y|Y|Y|Y|Y|}
\hline
\textbf{Model} & \textbf{Architecture} & \textbf{Model Size} & \textbf{Pre-training Tasks} & \textbf{Fine-tuning \& Zero-shot Abilities} & \textbf{Strengths} & \textbf{Weaknesses} \\
\hline
BERT-Large & Transformer (bidirectional) & 340M parameters & Masked Language Model (MLM), Next Sentence Prediction (NSP) & Effective fine-tuning; limited zero-shot & Deep bidirectional context, high accuracy & High computational cost \\
\hline
GPT-4 & Transformer (decoder-focused) & Significantly larger than GPT-3 & Unsupervised language modeling & Strong fine-tuning and zero-shot learning & Excellent text generation, wide task adaptability & Potential bias in generation, high computational cost \\
\hline
T5-11B & Transformer (encoder-decoder) & 11B parameters & Text-to-text framework & High fine-tuning ability; strong zero-shot learning & Unified framework for multiple tasks, scalability & Significant computational resources needed \\
\hline
BART-Large & Transformer (bidirectional encoder, autoregressive decoder) & Similar to BERT and GPT, exact count not specified & Denoising autoencoder & Effective in both text generation and comprehension & Flexible with noise, strong in generation \& comprehension tasks & High computational resources, potential limitations with structured language \\
\hline
\end{tabularx}
\caption{Comparison of various Pre-trained Large Language Models (the selected versions have the highest number of parameters available)}
\label{tab:lang_models}
\end{table*}


\subsubsection{BERT}

BERT (Bidirectional Encoder Representations from Transformers), introduced by Devlin et al. \cite{devlin2018bert}, is a pre-trained language model that has achieved state-of-the-art results in various natural language processing tasks. A key aspect of BERT's training process is the use of a Masked Language Model (MLM) objective. In this procedure, a predetermined portion of the sentence's input tokens is chosen to be masked or hidden from the model during training. Using the context that the other (non-masked words in the sentence) provide, the model is then trained to predict the original value of the masked words. BERT differs significantly from traditional unidirectional language models because it has a bidirectional understanding of context. BERT's strong contextual understanding and transfer learning capabilities give it a strong foundation for adapting to the abstractive text summarization task even though it was not designed specifically for it. In abstractive text summarization, BERT is used as an encoder to extract contextual information from the input text. The model is able to comprehend the intricate relationships between words and their meanings, which is essential to generate summaries that are accurate and coherent. This is made achievable by pre-training bidirectional transformers using the MLM training regimen. Its ability to capture both local contexts and long-range dependencies gives BERT a benefit over traditional sequence-to-sequence models.

Rothe et al. \cite{rothe2020leveraging} proposed a two-step strategy for abstractive text summarization by harnessing BERT's capacity to understand context-oriented details. Leveraging the contextual embeddings of an already trained BERT, they applied extractive techniques on a large corpus, permitting the model to grasp the structure and semantics of the source text and successfully extract salient information. This initial extractive step resulted in summaries that were more accurate and coherent. The results of this step are then fed into the abstractive summarization model, which centers around generating summaries that convey the main ideas while keeping up with text coherence. The benefits of extractive and abstractive summarization are integrated with this strategy, generating quality summaries with expanded comprehensibility and informativeness.

Dong et al. \cite{dong2019unified} presented a study on a unified language model, UniLM, for Natural Language Understanding (NLU) and Natural Language Generation (NLG) tasks. Utilizing shared knowledge between the two types of tasks is made possible by this unified approach, potentially enhancing performance in a variety of applications. Similar to BERT, UniLM is pre-trained using a variety of training objectives and is based on the transformer architecture. These objectives include masked language modeling (as in BERT), unidirectional (left-to-right or right-to-left) language modeling, and a novel Seq2Seq language modeling objective. UniLM can learn to comprehend and generate text by combining these objectives, making it suitable for NLU and NLG tasks. The CNN/Daily Mail dataset was one of the benchmark datasets used by the authors to evaluate UniLM for abstractive text summarization tasks. The experiment's findings demonstrated that the unified pre-training method significantly boosts performance when compared to other cutting-edge approaches.

Song et al. \cite{song2019mass} presented the Masked Sequence to Sequence Pre-training (MASS) method, a useful technique to enhance the ability of models to generate abstractive summaries. The MASS technique, which is based on the BERT architecture, employs the masked language model objective, which allows the model to learn contextual information from both input and output sequences. The authors developed the MASS approach primarily for tasks that required language generation, like abstractive summarization. By pre-training the model with the masked Seq2Seq method, they hoped to increase its ability to generate coherent, semantically meaningful summaries. In this technique, the input sequence is partially masked, and the model is trained to predict the masked tokens based on the unmasked ones. With the help of this pre-training technique, the model is better able to capture the context and structure of the input text, which is essential for generating accurate abstractive summaries.

\subsubsection{GPT}


Radford et al. \cite{radford2018improving, radford2019language} introduced the concept of Generative Pre-trained Transformers (GPT), a series of powerful language models designed for natural language understanding and generation. Unlike BERT, which is trained in a masked language model fashion where certain words in a sentence are hidden and predicted, GPT is trained using a generative approach. Specifically, it predicts the next word in a sequence given all previous words. Based on the transformer architecture, GPT and its successor, GPT-2, use unsupervised learning with a generative pre-training phase and fine-tuning on particular tasks. GPT-3, released in 2020, further scaled up the GPT approach to achieve strong performance on natural language tasks \cite{brown2020language}. OpenAI released GPT-3.5\footnote{https://en.wikipedia.org/wiki/GPT-3\#GPT-3.5} in 2022 by increasing the model size and training it on a larger dataset. Most recently, OpenAI released GPT-4\footnote{https://openai.com/research/gpt-4} in 2023, which is a significant improvement as compared to GPT-3.5 and also considers safety and ethics. Although the GPT models have shown excellent potential in a number of natural language processing tasks, including summarization, these works do not specifically focus on abstractive text summarization. Researchers have improved GPT models for abstractive text summarization tasks by employing the use of their generative nature, demonstrating their ability to generate coherent and contextually relevant summaries.

Zhu et al. \cite{zhu2022chinese} fine-tuned GPT-2 for abstractive text summarization in Chinese, a language that had not been extensively studied in relation to GPT-2's performance. They used a dataset of Chinese news articles and their summaries, leveraging the self-attention mechanisms and token-based representations of the transformer architecture to modify the GPT-2 model. Its performance was compared to baseline models, including Seq2Seq and Pointer-Generator Networks on a Chinese news summarization task. The authors found that the improved GPT-2 model surpassed the baselines in terms of ROUGE scores. This research underscores the importance of fine-tuning for language and domain-specific tasks, advancing the understanding of GPT-2's capacity for abstractive text summarization in non-English languages and enhancing its application in multilingual contexts.

The effectiveness of utilizing BERT and GPT-2 models for abstractive text summarization in the area of COVID-19 medical research papers was examined by Kieuvongngam et al. \cite{kieuvongngam2020automatic}, who used a two-stage approach to generate abstractive summaries. First, they employed a BERT-based extractive summarization model to select the most relevant sentences from a research article. In the second stage, the authors used the extracted sentences as input for the GPT-2 model, which generated abstractive summaries. The authors sought to generate summaries that are more informative and coherent by combining the benefits of extractive summarization provided by BERT and abstractive summarization provided by GPT-2. By contrasting the automatically generated summaries with human-written summaries, the researchers evaluated their method using a collection of research papers from the COVID-19 dataset. To evaluate the effectiveness of the generated summaries, they employed ROUGE metrics and human evaluation. The results show the potential of integrating BERT and GPT-2, as their two-stage strategy surpassed other cutting-edge techniques for automatic text summarization.

Alexandr et al. \cite{alexandr2021fine} fine-tuned GPT-3 for abstractive text summarization in the Russian language. They created a corpus of Russian news stories and accompanying summaries using a saliency-based summarization technique, which was subsequently used as input for GPT-3. The fine-tuned GPT-3 model's performance was compared to baseline models such as BERT, GPT-2, and the original GPT-3. They found that, as per the ROUGE metric, the improved GPT-3 model outperformed the baselines, emphasizing the significance of considering the unique challenges of different languages when adapting state-of-the-art models.

Bhaskar et al. \cite{bhaskar2023prompted} showcased GPT-3.5's ability in opinion summarization by summarizing a large collection of user reviews using methods like recursive summarization, supervised clustering, and extraction. They tested on two datasets: SPACE \cite{amplayo2021aspect} (hotel reviews) and FewSum \cite{bravzinskas2020few} (Amazon and Yelp reviews), with GPT-3.5 receiving high marks in human evaluations. However, standard metrics like ROUGE were found lacking in capturing summary nuances. The GPT-3.5 abstractive summaries were fluent but sometimes deviated from the original content or were over-generalized. To counter this, new metrics for faithfulness, factuality, and genericity were introduced. The study also examined the effects of pre-summarization and found that while GPT-3.5 was effective for shorter inputs, its accuracy decreased for longer reviews. Techniques like QFSumm \cite{ahuja2021aspectnews} helped in brevity but made summaries more generic. The team proposed topic clustering to enhance relevance, albeit with minor trade-offs.

Due to the public recent availability of GPT-3.5/4 and their productization as ChatGPT\footnote{https://chat.openai.com/}, their application in text summarization has become extensive. However, a significant limitation is that LLMs inherently cannot verify the accuracy of the information they generate. Addressing this, Chen et al. \cite{chen2023enhancing} introduced a novel approach to abstractive summarization that aim to overcome the truth comprehension challenges of LLMs. This method integrates extracted knowledge graph data and structured semantics to guide summarization. Building on BART, a leading
sequence-to-sequence pre-trained LLM, the study developed multi-source transformer modules as encoders, adept at handling both textual and graphical data. Decoding leverages this enriched encoding, aiming to improve summary quality. For evaluation, the Wiki-Sum dataset\footnote{https://paperswithcode.com/dataset/wikisum} is utilized. When compared to baseline models, results underscore the effectiveness of this approach in generating concise and relevant summaries.

Zhang et al. introduced \cite{zhang2023summit} a novel framework named SummIt, which is grounded on LLMs, particularly ChatGPT. Unlike conventional abstractive summarization techniques, SummIt adopts an iterative approach, refining summaries based on self-evaluation and feedback, a process reminiscent of human drafting and revising techniques. Notably, this framework circumvents the need for supervised training or reinforcement learning feedback. An added innovation is the integration of knowledge and topic extractors, aiming to augment the faithfulness and controllability of the abstracted summaries. Evaluative studies on benchmark datasets indicate superior performance of this iterative method over conventional one-shot LLM systems in abstractive tasks. However, human evaluations have pointed out a potential bias in the model, favoring its internal evaluation criteria over human judgment. This limitation suggests a potential avenue for improvement, possibly through incorporating human-in-the-loop feedback. Such insights are pivotal for future research focused on enhancing the efficacy of abstractive summarization using LLMs.

\subsubsection{T5}
Text-to-Text Transfer Transformer (T5) is a unified text-to-text transformer model that was developed by Raffel et al. \cite{raffel2020exploring} to handle a variety of NLP tasks, including abstractive text summarization. Like BERT and GPT, the T5 model is based on transformer architecture and aims to simplify the process of adapting pre-trained models to various NLP tasks by casting all tasks as text-to-text problems. T5's training protocol is different from BERT and GPT. The authors trained T5 in two steps. First, a de-noising autoencoder framework is used to pre-train the model using a large unsupervised text corpus. Reconstructing the original text from corrupted input is a pre-training task that helps the model learn the structure, context, and semantics of the natural language. Second, by transforming each task into a text-to-text format, the pre-trained model is fine-tuned on task-specific supervised datasets, such as summarization datasets. The authors utilized two NLP benchmarks: CNN/Daily Mail and XSum \cite{narayan2018don}, to demonstrate T5's efficacy. On these benchmarks, T5 achieved state-of-the-art results showcasing its capabilities in the abstractive summarization domain. The study also investigated the impact of model size, pre-training data, and fine-tuning strategies on transfer learning performance, offering insightful information about the T5 model's scalability and adaptability.

The effectiveness of the T5 model for abstractive text summarization in the Turkish language is examined by Ay et al. \cite{ay2023turkish}, who fine-tuned the T5 model on a dataset of news articles and corresponding summaries. They customized the model to generate abstractive summaries in Turkish by making use of the T5 architecture's capabilities. The researchers evaluated the fine-tuned T5 model and compared its performance with baseline models, such as BERT, GPT-2, and PEGASUS \cite{zhang2020pegasus}. The findings showed that the fine-tuned T5 model outperforms the baseline models and achieves high ROUGE scores, proving its efficiency in Turkish text summarization. This study contributed to the understanding of state-of-the-art models like T5 for abstractive text summarization in languages other than English.

Garg et al. \cite{garg2021news} compared the performance of T5 and BART alongside a custom-built encoder-decoder model and another model developed through transfer learning from T5, using a dataset comprising over 80,000 news articles and their corresponding summaries. The findings from the study corroborate their hypothesis, demonstrating that T5 indeed outperforms BART, the transfer learning model, and the custom encoder-decoder model. This research enhanced the comprehension of the effective application of pre-trained transformer models, such as T5, for both abstractive and extractive text summarization tasks, particularly within the domain of news articles.

Guo et al. \cite{guo2021longt5} introduced LongT5, a model that analyzes the effects of simultaneously adjusting input length and model size. LongT5 integrates attention ideas from long-input transformers (Extended Transformer Construction \cite{ainslie2020etc}) and adopts pre-training strategies from summarization pre-training (Pre-training with Extracted Gap-sentences for Abstractive Summarization Sequence-to-sequence - PEGASUS) into the scalable T5 architecture. LongT5's main contributions feature a novel scalable attention mechanism known as Transient Global (TGlobal) attention. TGlobal attention emulates Extended Transformer Construction's local/global attention mechanism without the need for extra inputs. LongT5 also adopts a PEGASUS-style Principle Sentences Generation pre-training objective. This new attention mechanism is more effective and flexible because it can be used without requiring alterations to the model inputs. On a number of summarization and question-answering tasks, LongT5 outperformed the original T5 models and achieved cutting-edge results. To promote additional study and development, the authors have made their architecture, training code, and pre-trained model checkpoints publicly available. 

Elmadany et al. \cite{elmadany2022arat5} showcased the effectiveness of T5-style models for the Arabic language. The authors presented three robust Arabic-specific T5-style models and evaluated their performance using a novel benchmark for ARabic language GENeration (ARGEN), which includes a range of tasks, including abstractive text summarization. While there are numerous tasks in the ARGEN benchmark, the emphasis on abstractive text summarization demonstrates the model's capacity to generate concise and coherent summaries of Arabic text sources. On all ARGEN tasks, including abstractive summarization, the authors discovered that their models significantly outperformed the multilingual T5 model (mT5), setting new state-of-the-art results. The ability of T5-style models to cope effectively with languages with various dialects and intricate structures is demonstrated by the effectiveness of AraT5 model in abstractive text summarization for the Arabic language. The research contributed to the development of more powerful and efficient models for abstractive text summarization in Arabic and other languages, highlighting the importance of creating language-specific models and benchmarks in natural language processing tasks, such as abstractive text summarization.

Zolotareva et al. \cite{zolotareva2020abstractive} compared the performance of T5 model with attention-based Seq2Seq models for abstractive text summarization. The authors concluded that the T5 model is effective in abstractive document summarization. They suggested that future research should explore the application of the Transformer method for multi-document summarization and test the T5 approach on other benchmark datasets.

\subsubsection{BART}

Lewis et al. \cite{lewis2019bart} presented BART (Bidirectional and Auto-Regressive Transformers), a denoising Seq2Seq pre-training approach suitable for tasks such as natural language generation, translation, comprehension, and abstractive text summarization. Using the transformer architecture, BART is trained by reconstructing original texts from their corrupted versions. This corruption is introduced through strategies like token masking, token deletion, and text shuffling. Unlike T5, which views every NLP task as a text-to-text problem and pre-trains with a ``fill-in-the-blank" task, BART adopts a denoising objective, aiming to restore corrupted text. This approach equips BART to handle tasks that demand understanding and reconstructing sentence structures. After this pre-training phase, BART can be fine-tuned on task-specific datasets, demonstrating its prowess in domains like abstractive text summarization. Notably, on the CNN/Daily Mail and XSum summarization benchmarks, BART surpassed prior models, underscoring its efficacy in the abstractive summarization domain.

Venkataramana et al. \cite{venkataramana2022abstractive} addressed the problem of abstractive text summarization and aimed to generate a concise and fluent summary of a longer document that preserves its meaning and salient points. The authors used BART, which is fine-tuned on various summarization datasets to adapt to different domains and styles of input texts. They also introduced an attention mechanism in BART’s layers, which allows the model to focus on the most relevant parts of the input text and avoid repetition and redundancy in the output summary. The authors evaluated BART on several benchmark datasets and compared it with other state-of-the-art models such as RoBERTa \cite{liu2019roberta}, T5, and BERT in terms of ROUGE scores, human ratings, and qualitative analysis. The paper demonstrated that BART is a powerful and versatile model for abstractive text summarization tasks, capable of generating high-quality summaries that are coherent, informative, and faithful to the original text.

Yadav et al. \cite{yadav2023fine} discussed enhancement of abstractive summarization by fine-tuning the BART architecture, resulting in a marked improvement in overall summarization quality. Notably, the adoption of Sortish sampling has rendered the model both smoother and faster, while the incorporation of weight decay has augmented performance by introducing model regularization. BartTokenizerFast, employed for tokenization, further refined the input data quality. Comparative analyses with prior models underscore the efficacy of the proposed optimization strategy, with evaluations rooted in the ROUGE score. 

La Quatra et al. \cite{la2022bart} introduced BART-IT, a Seq2Seq model grounded in the BART architecture, meticulously tailored for the Italian language. BART-IT is pre-trained on an expansive corpus of Italian texts, enabling it to capture language-specific nuances, and subsequently fine-tuned on benchmark datasets for abstractive summarization. The paper also discussed the ethical considerations surrounding abstractive summarization models, emphasizing the importance of responsible application.

Vivek et al. \cite{vivek2022sumbart} presented SumBART, an improved BART model for abstractive text summarization. Addressing BART's factual inconsistencies, three modifications were made to SumBART, resulting in better ROUGE scores and more accurate summaries. Evaluations on the CNN/Daily-mail and XSum datasets showed SumBART's summaries were more human-like than BART's.


\subsection{ Reinforcement Learning (RL) Approaches}

\begin{figure*}[htbp]
    \centering
    \includegraphics[width=\linewidth, frame]{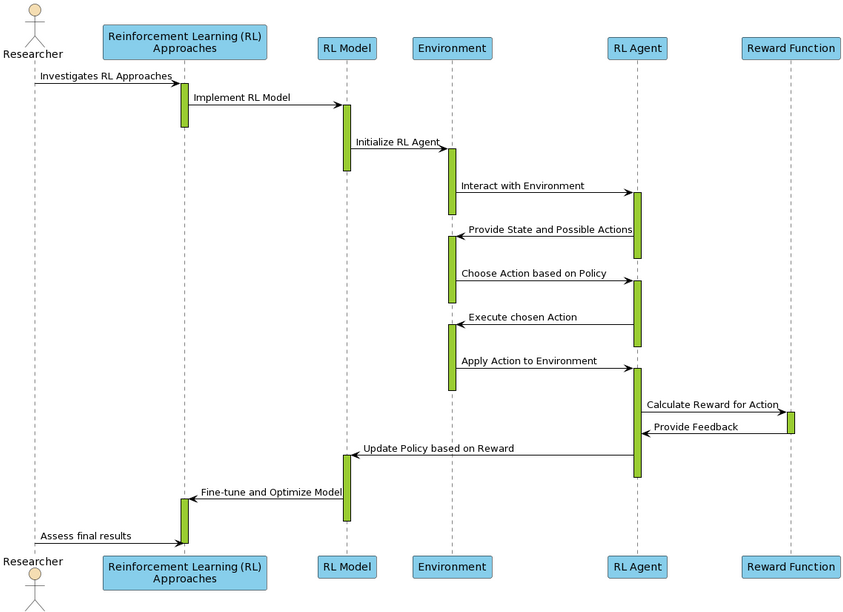}
    \caption{Reinforcement Learning approaches flow for abstractive summarization}
    \label{fig: RL}
\end{figure*}

Reinforcement learning (RL) is a type of machine learning where an agent interacts with the environment and learns to make optimal decisions by receiving rewards or penalties for its actions \cite{sutton2018reinforcement}. RL methods can be used for abstractive text summarization, where the model learns to generate concise summaries of documents by being rewarded for coherence, accuracy, and brevity \cite{ryang2012framework}. The sequence diagram in Figure \ref{fig: RL} illustrates a researcher utilizing reinforcement learning approaches in abstractive text summarization, where the model interacts with an environment to receive rewards or penalties, guiding it to generate optimal summaries. To offer an in-depth examination of the diverse techniques deployed in RL for abstractive text summarization, we have categorized them into two sub-classes: Policy Optimization Methods and Reinforcement Learning with Semantic and Transfer Learning Techniques. Policy Optimization Methods focus on acquiring the most effective strategy that directs the agent in making optimal choices while interacting with its environment, usually with the purpose of generating concise, accurate, and coherent summaries. Reinforcement Learning with Semantic and Transfer Learning Techniques combines RL with semantic analysis and transfer learning, giving the model the ability to understand contextual significance and apply knowledge from one domain to another in addition to making optimal decisions. Table \ref{tab:rl_summary_models} shows a comparison of Reinforcement Learning (RL) Approaches for Abstractive Text Summarization.

\renewcommand{\arraystretch}{0.8} 
\renewcommand\tabularxcolumn[1]{m{#1}} 

\begin{table*}
\centering
\begin{tabularx}{\textwidth}{|Y|Y|Y|Y|Y|Y|Y|}
\hline
\textbf{Authors and  Reference} & \textbf{Model or Framework} & \textbf{State Representation} & \textbf{Action} & \textbf{Reward} & \textbf{Policy Optimization} & \textbf{Additional Features} \\
\hline
Li et al. \cite{li2018actor} & Actor-Critic RL Framework & Document and partially generated summary & Token generation by Seq2Seq network & Quality of generated summary (Critic I and II) & Alternating training of actor and critic models & Attention-based Seq2Seq, evaluation with ROUGE metrics \\
\hline
Paulus et al. \cite{paulus2017deep} & RL-based Abstractive Summarization & Document and partially generated summary & Token generation or pointer mechanism & Maximizing ROUGE scores; Enhancing readability & Intra-attention mechanism, hybrid learning objective combining MLE loss with RL & Addressing long input sequences, focus on readability \\
\hline
Chen et al. \cite{chen2018fast} & Hybrid Extractive-Abstractive Model & Extracted sentences; Partially rewritten summary & Sentence selection; Rewriting by abstractor network & Sentence-level metric rewards, linguistic fluency & Actor-Critic policy gradient & Coarse-to-fine approach, speed optimization, novel n-grams generation \\
\hline
Celikyilmaz et al. \cite{celikyilmaz2018deep} & Deep Communicating Agents & Subsections of input text & Coordination among multiple agents for encoding & MLE, Semantic cohesion, RL loss & Multiple encoders with a single decoder & End-to-end training, intermediate rewards for unique sentence creation \\
\hline
Hyun et al. \cite{hyun2022generating} & MSRP & Document and partially generated summary of arbitrary lengths & Token generation by T5 transformer & Content preservation, Fluency, Length & Proximal Policy Optimization with adaptive clipping & Unsupervised method, multi-Summary based RL, SentenceBERT embeddings for content preservation \\
\hline
Jang et al. \cite{jang2021reinforced} & RL with ROUGE-SIM and ROUGE-WMD & Document and partially generated summary & Token generation & Semantic closeness to source, novelty, reduced direct copying & Decoding method based on ROUGE-SIM and ROUGE-WMD & Integration of Cosine Similarity and Word Mover's Distance with ROUGE-L, focus on grammatical integrity \\
\hline
Keneshloo et al. \cite{keneshloo2019deep} & Pointer-Generator with RL & Document and partially generated summary & Token generation & Negative expected reward, alignment with metrics like ROUGE & Self-critical policy gradient, Transfer RL approach & Addressing exposure bias, balancing training on source and target datasets \\
\hline
Wang et al. \cite{wang2018reinforced} & Topic-aware ConvS2S with RL & Document and partially generated summary & Token generation & Maximizing non-differentiable ROUGE metric & Self-Critical Sequence Training (SCST) & Topic-aware attention mechanism, addressing exposure bias through SCST \\
\hline
\end{tabularx}
\caption{Comparison of Reinforcement Learning (RL) Approaches for Abstractive Text Summarization}
\label{tab:rl_summary_models}
\end{table*}

\subsubsection{Policy Optimization Methods}

Li et al.\cite{li2018actor} presented an actor-critic \cite{grondman2012survey} RL training framework for enhancing neural abstractive summarization. The authors proposed a maximum likelihood estimator (MLE) and a global summary quality estimator in the critic part, and an attention-based Seq2Seq network in the actor component. The main contribution was an alternating training strategy to jointly train the actor and critic models. The actor generated summaries using the attention-based Seq2Seq network, and the critic assessed their quality using Critic I (MLE) and Critic II (a global summary quality estimator). Using the ROUGE metrics, the paper evaluated the proposed framework on three benchmark datasets: Gigaword, DUC-2004, and LCSTS \cite{hu2015lcsts} and achieved state-of-the-art results.

Paulus et al. \cite{paulus2017deep} presented an abstractive text summarization model that used RL to enhance performance and readability, especially for long input sequences. The model employed an intra-attention mechanism to avoid redundancy and utilized a hybrid learning objective combining maximum likelihood loss with RL. The model additionally included a softmax layer for token generation or a pointer mechanism to copy rare or unseen input sequence words. The approach optimized the model for the ROUGE scores without compromising clarity and showed cutting-edge results on the CNN/Daily Mail dataset, improving summary readability. The paper also highlights the limitations of relying solely on ROUGE scores for evaluation.

Chen et al. \cite{chen2018fast} proposed a hybrid extractive-abstractive model that adopts a coarse-to-fine approach inspired by humans, first selecting critical sentences using an extractor agent and then abstractly rewriting them through an abstractor network. The model also used an actor-critic policy gradient with sentence-level metric rewards to bridge the non-differentiable computation between the two neural networks while maintaining linguistic fluency. This integration was achieved using policy gradient techniques and RL. The system was optimized to reduce redundancy. On the CNN/Daily Mail dataset and the DUC2002 dataset used only for testing, the model generated state-of-the-art outcomes with significantly higher scores for abstractiveness. The model significantly improved training and decoding speed over earlier models and yielded summaries with a higher proportion of novel n-grams, a measure of greater abstractiveness. The model outperformed earlier available extractive and abstractive summarization algorithms in terms of ROUGE scores, human evaluation, and abstractiveness scores.

Celikyilmaz et al. \cite{celikyilmaz2018deep} generated abstractive summaries for lengthy documents by utilizing deep communicating agents within an encoder-decoder architecture. Multiple working agents, each responsible for a subsection of the input text, collaborate to complete the encoding operation. For the purpose of generating a targeted and comprehensive summary, these encoders are coupled with a single decoder that has been end-to-end trained using RL. In comparison to a single encoder or multiple non-communicating encoders, the results showed that multiple communicating encoders generate summaries of higher quality. Maximum likelihood (MLE), semantic cohesion, and RL loss were optimized during the training. Intermediate rewards, based on differential ROUGE measures, were incorporated to encourage unique sentence creation. Experiments conducted on the CNN/DailyMail and New York Times (NYT) \cite{sandhaus2008new} datasets showed better ROUGE scores compared to baseline MLE data. Human evaluations favored the summaries generated by deep communicating agents.

Hyun et al. \cite{hyun2022generating} introduced an unsupervised method for generating abstractive text summaries of varying lengths using RL. Their approach, Multi-Summary based Reinforcement Learning with Pre-training (MSRP), tackled the issue of generating summaries of arbitrary lengths while ensuring content consistency and fluency. MSRP's reward function is comprised of three components: content preservation, fluency, and length. The content preservation reward was determined using sentence similarity through SentenceBERT embeddings \cite{reimers2019sentence}. Fluency was gauged using a pre-trained GPT-2 model, and the length reward was based on the comparison of the generated summary's length to the target length. They employed a training method using Proximal Policy Optimization (PPO) \cite{schulman2017proximal} with an adaptive clipping algorithm, enabling quicker convergence and stability. The T5 transformer model served as the policy for training MSRP. When evaluated on the Gigaword dataset \cite{graff2003english}, MSRP consistently outperformed other unsupervised summarization models in terms of ROUGE scores. Despite using a larger model and an autoregressive approach, MSRP's inference time remained competitive due to its reward-based training and beam search implementation during summary creation.

\subsubsection{Reinforcement Learning with Semantic and Transfer Learning Techniques}

Jang et al. \cite{jang2021reinforced} addressed the limitations of traditional ROUGE-L based methods, which often generate repetitive summaries, by introducing two new RL reward functions: ROUGE-SIM and ROUGE-WMD. These functions integrate Cosine Similarity and Word Mover's Distance \cite{kusner2015word} respectively, with the ROUGE-L score, ensuring summaries are semantically close to the source text while promoting novelty and reducing direct copying. To generate accurate, innovative, and grammatically sound summaries, a decoding method based on these reward functions was proposed. The Gigaword dataset was employed to evaluate the new methods. The authors used three metrics: ROUGE-PACKAGE, novel n-grams, and Grammarly \footnote{https://www.grammarly.com} to measure summary quality, originality, and grammatical integrity. The proposed models surpassed various baselines, offering consistent performance, higher semantic value, improved abstractiveness, and reduced grammatical errors.

Keneshloo et al. \cite{keneshloo2019deep} introduced an RL framework for abstractive text summarization to address exposure bias and improve generalization to unfamiliar datasets. Exposure bias means that during the training phase, the model is provided with an accurate input at every decoder step, while in the testing phase, it must generate the next token based on its own output \cite{keneshloo2019deep}. They used a Pointer-Generator \cite{46111} model combined with an RL objective and a transfer reinforcement learning approach. Traditional methods relying on cross-entropy (CE) loss often suffer from exposure bias and poor generalization. In contrast, the proposed RL objective minimizes the negative expected reward, enabling the model to train based on its own output, in line with metrics like ROUGE. This self-critical policy gradient approach emphasizes better-performing samples during training. However, the model faced challenges in transfer learning due to training on the source dataset distribution. To counter this, they introduced a transfer RL approach using a shared encoder and a trade-off parameter to balance source and target dataset training. The authors used four datasets: Newsroom \cite{grusky2018newsroom}, CNN/Daily Mail, DUC 2003\footnote{https://www-nlpir.nist.gov/projects/duc/guidelines/2003.html}, and DUC 2004\footnote{https://www-nlpir.nist.gov/projects/duc/guidelines/2004.html}, for training and evaluating text summarization models. They evaluated the models using ROUGE-1, ROUGE-2, and ROUGE-L F1 scores. The proposed model achieved the best performance, outperforming baselines and state-of-the-art methods in generalizing to unseen datasets.

Wang et al. \cite{wang2018reinforced} presented a topic-aware Convolutional Seq2Seq (ConvS2S) model for abstractive text summarization, enhanced with RL. They introduced a new topic-aware attention mechanism that incorporates high-level contextual information, improving summarization. Their main contribution was using Self-Critical Sequence Training (SCST) \cite{rennie2017self} to address exposure bias, which arises when models train on ground-truth data distribution rather than their own. This bias often degrades test performance due to accumulated errors and inflexibility in generating summaries. To combat these issues, the authors applied SCST, a policy gradient algorithm that maximizes the non-differentiable ROUGE metric. This method allowed the model to learn its own distribution and optimize the evaluation measure. Using SCST, the model was incentivized to generate sequences with high ROUGE scores, mitigating exposure bias and enhancing test results. Three datasets, Gigaword, DUC-2004 and LCSTS, were used for evaluation. The proposed model achieved the best performance, getting the highest ROUGE-1, ROUGE-2 and ROUGE-L scores on Gigaword and LCSTS. It also had the top ROUGE-1 and ROUGE-L scores on DUC-2004.


\subsection{Hierarchical Approaches}

\begin{figure*}[htbp]
    \centering
    \includegraphics[width=\linewidth, frame]{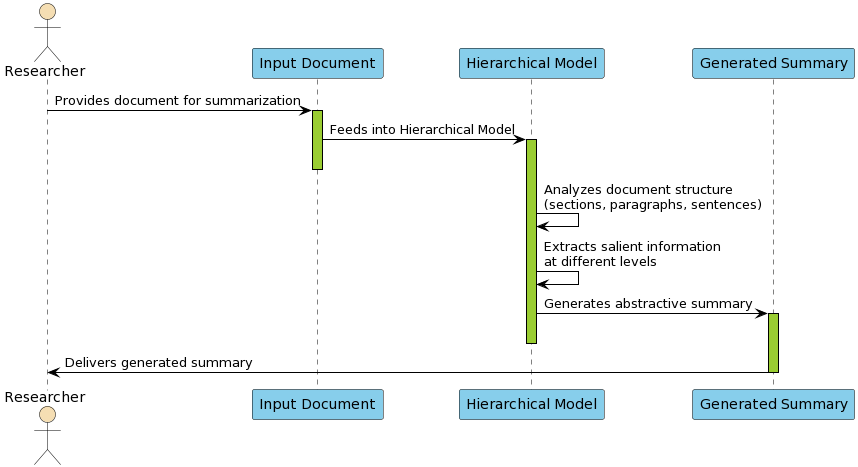}
    \caption{Hierarchical approaches flow for abstractive summarization}
    \label{fig: Hie}
\end{figure*}

Hierarchical approaches involve breaking down tasks into sub-problems and combining solutions to these sub-problems. Abstractive text summarization poses several challenges, such as handling long and complex documents, preserving the semantic and syntactic structure of the summary, and avoiding repetition and redundancy. To address these challenges, researchers have proposed hierarchical approaches \cite{yang2016hierarchical, fabbri2019multi} that exploit the hierarchical structure of natural language and model the source text at different levels of granularity. The sequence diagram in Figure \ref{fig: Hie} depicts the process of a researcher employing hierarchical models in abstractive text summarization, where the document undergoes multiple hierarchical encoding and decoding layers to generate a summarized output. To provide a more comprehensive understanding of Hierarchical Approaches in the context of abstractive text summarization, we have further divided them into three sub-classes: hierarchical LSTM-based approaches, hierarchical graph and network-based methods, and hierarchical attention and human-like methods. This classification makes it possible to better analyze the various techniques utilized in Hierarchical Approaches for abstractive text summarization. Table \ref{tab:hierarchical_summary_models} a comparison of Hierarchical Models/Frameworks for Abstractive Text Summarization.

\renewcommand{\arraystretch}{1} 

\begin{table*}
\centering
\begin{tabularx}{\textwidth}{|Y|Y|Y|Y|Y|}
\hline
\textbf{Authors and Reference} & \textbf{Specific Focus} & \textbf{Hierarchical Components} & \textbf{Model/Framework Used} & \textbf{Unique Features} \\
\hline
Nguyen et al. \cite{nguyen2021hierarchical} & Two-level LSTM architecture & Token and Document Level LSTMs & Hierarchical LSTM Encoder-Decoder & Captures hierarchical structure of documents \\
\hline
Song et al. \cite{song2019abstractive} & Phrase-based summarization & Multiple Order Semantic Parsing (MOSP) & LSTM-CNN & Focus on fine-grained semantic phrases \\
\hline
Zhang et al. \cite{zhang2018hierarchical} & Handling rare phrases and deep semantic understanding & Multi-step attention mechanism, CopyNet & Seq2Seq model with hierarchical attention & Focus on relevance, coherence, and handling OOV words \\
\hline
Yang et al. \cite{yang2020hierarchical} & Emulating human reading cognition process & Knowledge-based hierarchical attention, multitask module, DD-GAN & HH-ATS & Focus on informativeness and fluency of summaries \\
\hline
Zhao et al. \cite{zhao2019abstractive} & Abstractive meeting summarization & Adaptive segmental encoder networks, reinforced decoder networks & Hierarchical adaptive segmental network learning framework & Utilizes structure of meeting conversations \\
\hline
Qiu et al. \cite{qiu2022abstractive} & Exploiting hierarchical structure of documents & Hierarchical Graph Neural Network (HierGNN) & HierGNN-PGN, HierGNN-BART & Focus on identifying salient information \\
\hline
\end{tabularx}
\caption{Comparison of Hierarchical Models/Frameworks for Abstractive Text Summarization}
\label{tab:hierarchical_summary_models}
\end{table*}

\subsubsection{Hierarchical LSTM-based Approaches}

Nguyen et al. \cite{nguyen2021hierarchical} presented a method for abstractive text summarization that used a hierarchical Long Short-Term Memory (LSTM) encoder-decoder model. The key contribution is the development of a two-level LSTM architecture, which successfully captures the hierarchical structure of documents and improves sentence and paragraph representation and understanding. The hierarchical encoder consists of two LSTM layers that collaborate to process and represent textual information. The first LSTM layer operates at the token level and captures the relationships among words within a sentence. This layer processes the individual words in the context of the words around them to generate sentence-level representations. The second LSTM layer, which acts at the document level, uses these sentence-level representations as inputs. This layer effectively captures the relationships among sentences within a document by processing the sentence-level representations sequentially to obtain document-level representations. The ROUGE-1, ROUGE-2, and ROUGE-L scores showed that the hierarchical LSTM encoder-decoder model outperformed strong baseline models on two benchmark datasets: Gigaword and Amazon reviews from Stanford Network Analysis Project (SNAP) \footnote{https://snap.stanford.edu/data/web-Amazon.html}.  

Song et al. \cite{song2019abstractive} proposed a novel LSTM-CNN based abstractive text summarization framework called ATSDL that generated summaries by exploring fine-grained semantic phrases rather than just sentences. The ATSDL framework had two main stages – phrase extraction from source sentences using a technique called Multiple Order Semantic Parsing (MOSP), and summary generation using a LSTM-CNN model that learned phrase collocations. MOSP extracted subject, relational and object phrases by scattering sentences into fragments and restructuring them into a binary tree, providing richer semantics than keywords. The LSTM-CNN model took phrase sequences as input, learning phrase collocations and capturing both semantics and syntactic structure, overcoming limitations of extractive and abstractive models. The model had a convolutional phrase encoder and recurrent decoder that could generate or copy phrases, handling the problem of rare words. Refining and combining similar phrases before LSTM-CNN training reduced phrase redundancy and sparsity, improving learning. Experiments on CNN/DailyMail datasets showed that the model outperformed state-of-the-art abstractive and extractive models in ROUGE scores. The generated summaries were composed of natural sentences meeting syntactic requirements, while capturing semantics – demonstrating benefits of the hierarchical phrase-based approach.

\subsubsection{Hierarchical Attention and Human-like Approaches}

Zhang et al. \cite{zhang2018hierarchical} employed the CopyNet \cite{gu2016incorporating} mechanism and the hierarchical attention model for abstractive summarization. The authors developed an approach based on handling rare phrases and deep semantic understanding. This Seq2Seq model is enhanced with a multi-step attention mechanism that enables it to generate summaries that are more relevant and coherent and reflect the compositional structure of the document. The model can also deal with OOV and rare words by copying them directly from the source text using the CopyNet approach. The authors used two datasets for the experimental evaluation: Gigaword Corpus and Large-Scale Chinese Short Text Summarization Dataset (LCSTS) Corpus. The experiments were conducted on three different models, namely Words-lvt2k-1sent (basic attention encoder-decoder model), Words-lvt2k-2sent (trained on two sentences at the source), and Words-lvt2k-2sent-hieratt (improves performance by learning the relative importance of the first two sentences). ROUGE-1, ROUGE-2, and ROUGE-L were used for evaluation. Results showed that the proposed hierarchical attention model demonstrated effectiveness and efficiency in generating summarizations for different types of text, outperforming the baseline models on the Gigaword Corpus. However, on the LCSTS Corpus, the hierarchical attention model did not outperform the baseline attentional decoder.

Yang et al. \cite{yang2020hierarchical} developed a method for abstractive text summarizing named Hierarchical Human-like Abstractive Text Summarization (HH-ATS). By incorporating three key elements—a knowledge-based hierarchical attention module, a multitask module, and a Dual Discriminative Generative Adversarial Networks (DD-GAN) framework—the model emulates the process of human reading cognition. These components correspond to the three phases of the human reading cognition process: rough reading, active reading, and post-editing. The knowledge-based hierarchical attention module in the HH-ATS model captures the global and local structures of the source document to focus on critical information. By simultaneously training the model on text categorization and syntax annotation tasks, the multitask module enhances the model’s performance. The DD-GAN framework refines the summary quality by introducing a generative model and two discriminative models that evaluate the informativeness and fluency of generated summaries. The authors experimented on CNN/Daily Mail and Gigaword datasets. ROUGE-1, ROUGE-2, and ROUGE-L were used for evaluation. A human evaluation was also conducted, in which the summaries generated by the HH-ATS model were compared to the best-performing baseline method. The results demonstrated that the strategy can generate summaries that were more informative and fluent. The performance of the HH-ATS model was examined using an ablation study, which revealed that all three of the components were essential to the model’s overall success. Additionally, the convergence analysis showed that the HH-ATS model converges more rapidly and stably than the baseline approaches, demonstrating the efficacy of the training strategy for the model.

\subsubsection{Hierarchical Graph and Network-based Approaches}

Using a hierarchical adaptive segmental network learning framework, Zhao et al. \cite{zhao2019abstractive} developed method for addressing the issue of abstractive meeting summarization. The two key elements of the method are adaptive segmental encoder networks for learning the semantic representation of the conversation contents and reinforced decoder networks for generating the natural language summaries. By adaptively segmenting the input text depending on conversation segmentation cues, the adaptive segmental encoder networks are created to take advantage of the structure of meeting conversations. A conversation segmentation network that recognizes segment boundaries and provides this information to the encoder is used to identify these cues. This approach enables the encoder to learn the semantic representation of meeting conversations while considering their inherent structure, which is crucial for generating accurate summaries. The reinforced decoder network is based on segment-level LSTM networks. The authors used AMI meeting corpus \cite{carletta2005ami} containing 142 meeting records and evaluated the results using ROUGE metrics.The authors identified that the maximum likelihood estimation used for training the decoder network can lead to suboptimal performance, and used an RL framework to train the decoder network. The methods, HAS-ML (trained with maximum likelihood estimation) and HAS-RL (trained with reinforcement learning), achieve high performance, indicating the effectiveness of the hierarchical adaptive segmental network learning framework. The HAS-RL method performs better than HAS-ML, which shows the effectiveness of RL in generating summaries in the problem of abstractive meeting summarization.

Qiu et al. \cite{qiu2022abstractive} presented a Hierarchical Graph Neural Network (HierGNN) approach to improve abstractive text summarization by exploiting the hierarchical structure of input documents using three steps: 1) learning a hierarchical document structure by a latent structure tree learned via sparse matrix-tree computation; 2) propagating sentence information over this structure via a message-passing node propagation mechanism called Layer-Independent Reasoning (LIR) to identify salient information; and 3) concentrating the decoder on salient information via a graph-selection attention mechanism (GSA). HierGNN is used in two architectures: HierGNN-Pointer-Generator Network (HierGNN-PGN) and HierGNN-BART. To improve sentence representations, both models include a two-layer HierGNN on top of the sentence encoder. Experiments show that the HierGNN model outperforms strong LLMs like BART in average ROUGE-1/2/L for CNN/DailyMail and XSum. Furthermore, human evaluations show that the model’s summaries generated by HierGNN are more relevant and less redundant than the baselines. The authors also identified that HierGNN more effectively analyses long inputs and synthesizes summaries by fusing many source sentences instead of compressing a single source sentence. However, The HierGNN model is based on an inverted pyramid writing style, which may not be applicable to other sorts of input texts. In addition, the model’s complexity rises as a result of the HierGNN encoder.

\subsection{Multi-modal Summarization}

\begin{figure*}[htbp]
    \centering
    \includegraphics[width=\linewidth, frame]{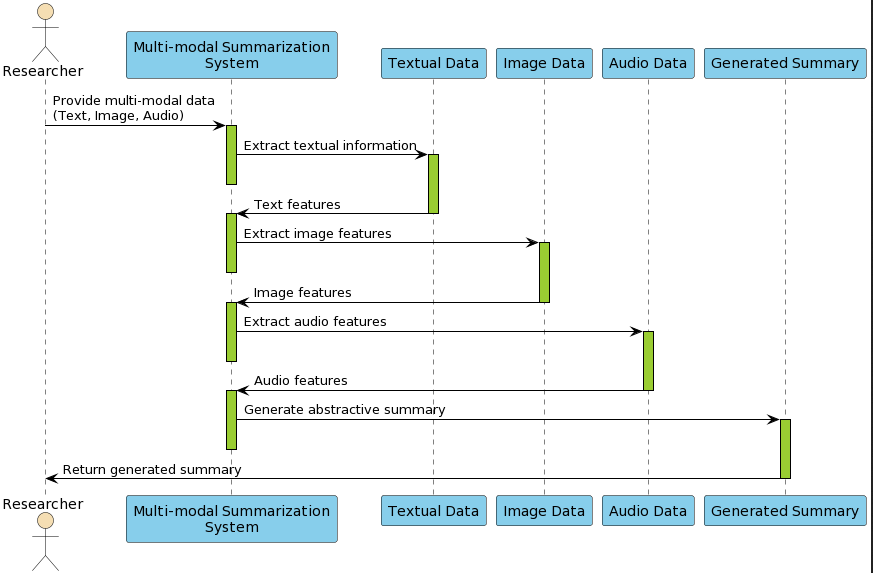}
    \caption{Multi-Modal flow for abstractive text summarization.}
    \label{fig: MM}
\end{figure*}

In many real-world scenarios, text documents are often accompanied by other modalities, such as images or videos, that provide complementary or supplementary information. Multi-modal summarization aims to leverage the information from multiple modalities to generate a coherent and comprehensive summary that reflects the salient aspects of the entirety of the input. The sequence diagram in Figure \ref{fig: MM} depicts a researcher employing multi-modal approaches in abstractive text summarization, integrating both textual and non-textual data sources to generate comprehensive and context-rich summaries. To provide a more comprehensive understanding of multi-modal approaches in the context of abstractive text summarization, we have further divided them into two sub-classes: Text-Image Summarization and Text-Video Summarization. Table \ref{tab:multimodal_summary_models} presents a comparison of Multi-modal Models/Frameworks for Abstractive Text Summarization.

\renewcommand{\arraystretch}{1}

\begin{table*}
\centering
\begin{tabularx}{\textwidth}{|Y|Y|Y|Y|Y|}
\hline
\textbf{Authors and Reference} & \textbf{Specific Focus} & \textbf{Modalities Considered} & \textbf{Model/Framework Used} & \textbf{Unique Features} \\
\hline
Chen et al. \cite{chen2018abstractive} & Multi-modal attention in summarization & Text, Images & Hierarchical Encoder-Decoder with Multi-modal Attention & Hierarchical attention for text and images, OOV replacement mechanism \\
\hline
Li et al. \cite{li2018multi} & Multi-modal sentence summarization & Text, Images & Seq2Seq with Hierarchical Attention & Modality-based attention, image filters for noise reduction \\
\hline
Zhu et al. \cite{zhu2018msmo} & Selecting pertinent images for summaries & Text, Images & Multimodal Attention Model & Multi-modal attention for text summary and image selection \\
\hline
Zhu et al. \cite{zhu2020multimodal} & Multi-modal summarization with output evaluation & Text, Images & Multimodal Summarization with Multimodal Output (MSMO) Model & Cross-Entropy loss for image selection, Negative Log-Likelihood loss for summary generation \\
\hline
Li et al. \cite{li2020aspect} & Aspect-aware summaries for e-commerce products & Text, Images & Aspect-aware Multimodal Summarization Model & Pointer-generator networks integrating visual data \\
\hline
Liu et al. \cite{liu2022abstractive} & Open-domain video text summarization & Text, Video & Multistage Fusion Network with Forget Gate (MFFG), Single-Stage Fusion Network with Forget Gate (SFFG) & Multistage and Single-Stage Fusion Networks \\
\hline
Khullar et al. \cite{khullar2020mast} & Leveraging audio, text, and video for summarization & Audio, Text, Video & Seq2Seq Model with Trimodal Attention & Trimodal Hierarchical Attention layer \\
\hline
Raji et al. \cite{raji2022abstractive} & Summarization from text, image, and video data & Text, Image, Video & LSTM Attention Encoder-Decoder Model & Integration of Tesseract-OCR for image text extraction \\
\hline
Fu et al. \cite{fu2020multi} & Integrating text and video summarization & Text, Video & Multi-Modal Summarization Model with Bi-Hop Attention & Bi-hop attention mechanism, advanced late fusion method \\
\hline
Li et al. \cite{li2020vmsmo} & Video-based multimodal summarization & Text, Video & Video-based Multimodal Summarization with Multimodal Output (VMSMO) Model & Dual-Interaction-based Multimodal Summarizer (DIMS) \\
\hline
\end{tabularx}
\caption{Comparison of Multi-modal Models/Frameworks for Abstractive Text Summarization}
\label{tab:multimodal_summary_models}
\end{table*}

\subsubsection{Text-Image Summarization}

Chen et al. \cite{chen2018abstractive} developed a multi-modal attentional mechanism that pays attention to original sentences, images, and captions within a hierarchical encoder-decoder architecture. They extended the DailyMail dataset and introduced E-DailyMail corpus by extracting images and captions from HTML-formatted texts. During the encoding phase, a hierarchical bi-directional RNN using GRU is employed to encode the sentences and the text documents, while an RNN and a CNN are used to encode the image set. In the decoding stage, text and image encodings are combined as the initial state, and an attentional hierarchical decoder is used to generate the text summary while focusing on the original phrases, photos, and captions. To generate summaries, the authors propose a multi-modal beam search method. Beams scores are based on the bigram overlaps of the generated sentences and captions. Additionally, they created an OOV replacement mechanism, enhancing the effectiveness of summarization. The main evaluation metrics used were ROUGE scores. When compared to existing neural abstractive models, extractive models, and models without multi-modal attention, the model that attends to images performs significantly better. Furthermore, the experiments demonstrate that their model is capable of producing informative summaries of images.

Li et al. \cite{li2018multi} introduced a multi-modal sentence summarization task using a sentence-image pair. The authors constructed a multi-modal sentence summarization corpus that consists of 66,000 summary triples (sentence, image, summary). To effectively integrate visual elements, the authors developed image filters and proposed a modality-based attention mechanism to focus on image patches and text units separately. They designed a Seq2Seq model with hierarchical attention mechanisms that concentrated on both image and text details. Visual elements were incorporated into the model to initiate the target language decoder, and image filter modules were used to reduce visual noises. ROUGE metrics were used for evaluation. They discovered that initializing the decoder using images improved performance, indicating that image features effectively captured key points of source texts. The multi-modal model was found to be more abstractive than the text-only model. Moreover, the multi-modal coverage technique effectively reduced word repetition.

Zhu et al. \cite{zhu2018msmo} introduced a novel task in multimodal summarization whose objective is to choose the image most pertinent to the abstractive summary from the multimodal input. The authors developed this task in response to their findings that multimodal outputs considerably increase user satisfaction upto 12.4\% in terms of informativeness. The authors constructed a multi-modal dataset \footnote{http://www.nlpr.ia.ac.cn/cip/jjzhang.htm} from the DailyMail corpus. They developed a multimodal attention model, which can concurrently generate text summaries and choose the input’s most suitable image. The Multimodal Automatic Evaluation (MMAE) method considers both intra-modality salience and inter-modality relevance in order to evaluate the multimodal outputs. A text encoder, an image encoder, a multimodal attention layer, and a summary decoder are the four main parts of the model. During the decoding process, the multimodal attention layer combines textual and visual information. A unidirectional LSTM used in the summary decoder generates the text summary and selects the most relevant image based on the visual coverage vector. Their multimodal attention model achieved better MMAE scores compared to extractive methods.

For Multimodal Summarization with Multimodal Output (MSMO), Zhu et al. \cite{zhu2020multimodal} used a multimodal objective function that combines the Cross-Entropy loss (CE) for image selection and the Negative Log-Likelihood loss (NLL) for summary generation. The MSMO dataset, which consists of online news stories combined with several image-caption pairs and multi-sentence summaries, is used by the authors in their experiments. ROUGE-ranking and Order-ranking are two additional techniques the authors presented for converting a text reference into a multimodal reference. They also developed an evaluation metric built on joint multimodal representation, which projects the model output and multimodal reference into a joint semantic space. Experiment results show that the model performs well in terms of both automatic and manual evaluation metrics, and has a better correlation with human judgments.

Li et al. \cite{li2020aspect} generated abstractive summaries for Chinese e-commerce products that incorporate both visual and textual information. The aspect-aware multimodal summarization model efficiently combines visual data from product photos and highlights the most crucial features of a product that are valuable for potential consumers. The model, based on pointer-generator networks, integrates visual data using three different methods: initializing the encoder with the global visual feature, initializing the decoder with the global visual feature, and producing context representations with the local visual features. The authors created CEPSUM, a large-scale dataset for summarizing Chinese e-commerce products, with over 1.4 million summaries of products that were manually written together with comprehensive product data that includes a picture, a title, and additional textual descriptions. The Aspect Segmentation algorithm, a bootstrapping method that automatically expands aspect keywords, is used to mine aspect keywords. A number of text-based extractive and abstractive summarization techniques, LexRank \cite{erkan2004lexrank}, Seq2seq, Pointer-Generator, and MASS \cite{song2019mass}, were compared to the model using the ROUGE score and manual evaluations, with the finding that the aspect-aware multimodal pointer-generator model outperformed the compared techniques.

\subsubsection{Text-Video Summarization}
Liu et al. \cite{liu2022abstractive} introduced two models for multi-modal abstractive text summarization of open-domain videos: Multistage Fusion Network with Forget Gate (MFFG) and Single-Stage Fusion Network with Forget Gate (SFFG). MFFG integrated multi-source modalities such as video and text by utilizing a multistage fusion schema and a fusion forget gate module, enhancing the model's ability to generate coherent summaries. SFFG, a simplified version of MFFG, reduced model complexity by sharing features across stages and used the source input text to improve summary representation. The authors used ROUGE (1,2,L), BLEU(1,2,3,4) \cite{papineni2002bleu}, CIDEr \cite{vedantam2015cider}, and METEOR metrics to evaluate model performance. Results revealed that MFFG and SFFG outperformed other methods in terms of Informativeness and Fluency. Specifically, SFFG excelled with ground truth transcript data in the How2 and How2-300h datasets \cite{sanabria2018how2}, while MFFG demonstrated superior anti-noise capabilities using automatic speech recognition-output transcript data. 

Using data from three different modalities—audio, text, and video—Khullar et al. \cite{khullar2020mast} introduced a Seq2Seq model for multimodal abstractive text summarization. Earlier studies concentrated on textual and visual modes, overlooking the potential of audio data to help generate more accurate summaries. A Trimodal Hierarchical Attention layer is used to fully leverage all three modalities. The model’s capacity to generate coherent and comprehensive summaries is improved by this layer, which enables the model to selectively attend to the most pertinent information from each modality. This specialized attention layer integrates audio, text, and video data from independent encoders. The output derived from the attention layer is utilized as the input for the decoder, which generates the abstractive summary. The authors used ROUGE scores and a Content F1 metric \cite{palaskar2019multimodal} for evaluation and showed that their model outperforms baselines. The study demonstrates that the inclusion of audio modality and the attention layer's ability to effectively extract relevant information from multiple modalities significantly improves the performance of multimodal abstractive text summarization.

Raji et al. \cite{raji2022abstractive} developed an LSTM attention encoder-decoder model to generate abstract summaries from text, image, and video data. To process image data and extract text, the system leverages the Tesseract-OCR engine \cite{patel2012optical}. It extracts audio from video files before converting them to text. It employs an RNN encoder-decoder architecture with an attention mechanism and LSTM cells to improve its handling of long-range dependencies. By removing several encoded vectors from the source data and generating abstractive summaries, this methodology enhances the summarization procedure. For training and validation, the authors use the Amazon fine food dataset \footnote{https://www.kaggle.com/datasets/snap/amazon-fine-food-reviews}, which consists of reviews of foods, with the description serving as the input variable and the title serving as the target variable. The authors calculated F1, precision and recall scores for Rouge-1, Rouge-2, and Rouge-L. This approach performs well in a variety of applications, such as generating abstracts for lengthy documents or research papers and enhancing the overall accessibility and comprehension of information in various formats.

Fu et al. \cite{fu2020multi} developed a model that incorporates four modules: feature extraction, alignment, fusion, and bi-stream summarizing. They utilized a bi-hop attention mechanism to align features and an advanced late fusion method to integrate multi-modal data. A bi-stream summarization technique enabled simultaneous summarizing of text and video. The authors introduced the MM-AVS dataset, derived from Daily Mail and CNN websites, containing articles, videos, and reference summaries. They evaluated using ROUGE scores for text summarization and cosine image similarity for video summarization. The proposed model surpassed existing methods in multi-modal summarization tasks, emphasizing the effectiveness of the bi-hop attention, improved late fusion, and bi-stream summarization approaches.

Li et al. \cite{li2020vmsmo} introduced Video-based Multimodal Summarization with Multimodal Output (VMSMO) to automatically select a video cover frame and generate a text summary from multimedia news articles. Their model, the Dual-Interaction-based Multimodal Summarizer (DIMS), comprises three primary components: Dual Interaction Module, Multi-Generator, and Feature Encoder. This model conducts deep interactions between video segments and articles and subsequently generates both a written summary and a video cover decision. For this work, the authors compiled the first large-scale VMSMO dataset from Weibo\footnote{https://us.weibo.com/index}, China’s largest social network website, including videos with cover images and articles with text summaries. The model was evaluated using standard Rouge metrics for text summary and mean average precision (MAP) \cite{zhou2018multi} and recall at position (Rn@k) \cite{tao2019multi} for video cover frame selection. Rn@k measures whether the positive sample is ranked in the first k positions of n candidates. The DIMS model's performance was benchmarked against several baselines. Experimental results, based on both automatic evaluations and human judgments, indicated that DIMS outperformed state-of-the-art methods.

\section{Model Scalabity and Computational Complexity in Abstractive Summarization}

Table \ref{table:models_comparison} provides a concise comparative analysis of various models in abstractive text summarization, such as by highlighting their parameters, computational demands, and performance metrics. This comparison offers valuable insights into the strengths and limitations of each model type.

\subsection{Traditional Seq2Seq based Models}
\begin{itemize}
    \item Model Scales and Parameters: Traditional Seq2Seq models, which primarily use LSTM or GRU units, typically contain parameters ranging from tens to hundreds of millions. These models marked the early advancements in neural network-based text summarization by learning to map input sequences to output sequences \cite{sutskever2014sequence}.
    \item Computational Complexity and Resource Consumption: Seq2Seq models require a substantial amount of computational power for both training and inference, particularly when processing longer text sequences. They are less resource-intensive compared to LLMs but offer limited capabilities in handling complex contextual relationships in text \cite{yousuf2021systematic}.
    \item Comparative Analysis of Performance and Resource Efficiency: While offering moderate performance in summarization tasks, these models are generally faster and more resource-efficient, making them suitable for scenarios with computational constraints \cite{chiu2018state, joshi2023deepsumm}.
\end{itemize}

\subsection{Pre-trained Large Language Models}
\begin{itemize}
    \item Pre-trained models such as BERT, GPT-3, and GPT-4 have significantly redefined the scale of model parameters in the field. BERT's base model, for instance, contains around 110 million parameters, setting a new standard for deep learning models. Following this, GPT-3 pushed the boundaries further with an unprecedented 175 billion parameters. Most recently, GPT-4 has surpassed its predecessors, boasting a staggering 1760 billion parameters, representing the continuous evolution and rapid growth in the size and complexity of these models \cite{devlin2018bert, brown2020language, rosol2023evaluation}.
    \item Computational Complexity and Resource Consumption: The training of these models requires extensive computational resources, including high-performance GPUs and significant amounts of memory. Inference, while providing high-quality outputs, is resource-intensive, particularly for real-time applications \cite{hadi2023survey}.
    \item Comparative Analysis of Performance and Resource Efficiency: These models deliver state-of-the-art results in summarization, achieving high levels of accuracy, coherence, and fluency. However, their training and operational costs are substantial, making them less accessible for smaller-scale applications \cite{bai2024beyond}.
\end{itemize}

\subsection{Reinforcement Learning (RL) Approaches}
\begin{itemize}
    \item Model Scales and Parameters: RL-based models in text summarization vary in size, but their complexity lies more in the training process, which involves learning to optimize a reward function, often based on human feedback or specific performance metrics \cite{alomari2022deep}.
    \item Computational Complexity and Resource Consumption: RL models are computationally demanding, primarily during the training phase where they require numerous iterations to converge to an optimal policy. This makes them resource-intensive, both in terms of computational power and time \cite{zhan2020experience}.
    \item Comparative Analysis of Performance and Resource Efficiency: RL approaches can tailor summaries more closely to specific user preferences or criteria \cite{li2019deep} but at the cost of increased computational resources and training complexity \cite{chen2021adaptive}.
\end{itemize}

\subsection{Hierarchical Approaches}
\begin{itemize}
    \item Model Scales and Parameters: Hierarchical models, which process text at various levels such as sentence and paragraph levels, are characterized by a higher number of parameters due to their multi-layered nature. For instance, the Hierarchical Attention Network (HAN) \cite{yang2016hierarchical} model proposed by Yang et al. integrates two levels of attention mechanisms — one at the word level and another at the sentence level — which significantly increases the total parameter count of the model. This layered approach allows for a deeper understanding of the text structure but also results in larger model size, often encompassing millions of parameters to account for the complexity of processing and synthesizing information at different textual hierarchies \cite{li2023hierarchical}.
    \item Computational Complexity and Resource Consumption: These models necessitate substantial computational power for processing multiple layers of text information, often leading to increased training and inference times \cite{hazra2021sustainable}.
    \item Comparative Analysis of Performance and Resource Efficiency: They are effective in capturing the overall structure and meaning of large documents but require more computational resources compared to simpler models \cite{diao2020crhasum}.
\end{itemize}

\subsection{Multi-modal Summarization}
\begin{itemize}
    \item Model Scales and Parameters: Multi-modal summarization models integrate various data types (text, images, audio) and therefore have complex and large architectures to process and synthesize information from different modalities \cite{bang2023multitask}.
    \item Computational Complexity and Resource Consumption: Integrating multiple data types in multi-modal summarization models, like the one by Li et al. \cite{li2019keep}, significantly increases computational demands. These models require advanced processing power and extensive memory to analyze and synthesize text, images, and audio. This complexity in processing different modalities leads to higher resource requirements during both training and inference stages, making them more resource-intensive than unimodal models \cite{jangra2023survey}.
    \item Comparative Analysis of Performance and Resource Efficiency: Multi-modal summarization models enhance the richness of summaries by combining various data types, but this advantage comes with increased computational costs. As exemplified by Sun et al. \cite{sun2018multi}, while these models achieve higher accuracy and coherence by integrating text and visual information, they demand greater computational resources, including longer training periods and more memory. Balancing enhanced performance with resource efficiency is a critical aspect of developing multi-modal summarization models \cite{kaur2021comparative}.
\end{itemize}

\newcolumntype{Y}{>{\centering\arraybackslash}X}

\renewcommand{\arraystretch}{1.2} 
\renewcommand\tabularxcolumn[1]{m{#1}} 

\begin{table*}
\centering
\begin{tabularx}{\textwidth}{|Y|Y|Y|Y|Y|Y|Y|Y|Y|}
\hline
\textbf{Model Type} & \textbf{Parameters} & \textbf{Training Time} & \textbf{Inference Time} & \textbf{Accuracy} & \textbf{Coherence} & \textbf{Fluency} & \textbf{Resource Consumption} & \textbf{Reference} \\
\hline
Traditional Seq2Seq & 50M - 500M & 2 - 14 days & 1 - 5 seconds & Moderate & Moderate & Moderate & Moderate & \cite{sutskever2014sequence} \\
\hline
Pre-trained Large Models (BERT, GPT-3, GPT-4 etc.) & 110M (BERT) to 175B (GPT-3) to 1.76T (GPT-4) & BERT: 4 weeks, GPT-3: 2 months, GPT-4: 3 months & BERT: 2 seconds, GPT-3: 5 seconds, GPT-4: 5 seconds & High to Very High & High to Very High & High to Very High & High & \cite{devlin2018bert}, \cite{brown2020language}, GPT-4\footnote{https://openai.com/research/gpt-4} \\
\hline
RL Approaches & Varies with architecture & 3 - 6 weeks & 2 - 10 seconds & High & High & High & High & \cite{paulus2017deep} \\
\hline
Hierarchical Approaches & 100M - 1B & 4 - 8 weeks & 2 - 6 seconds & High & High & High & High & \cite{yang2016hierarchical} \\
\hline
Multi-modal Summarization & 200M - 2B (depending on modalities) & 5 - 10 weeks & 3 - 7 seconds & High & High & High & Very High & \cite{sun2018multi} \\
\hline
\end{tabularx}
\caption{Comparative Analysis of Different Types of Models in Abstractive Text Summarization}
\label{table:models_comparison}
\end{table*}

\section{Issues, Challenges, and Future Directions for Abstractive Summarization}

\begin{figure*}[htbp]
    \centering
    \includegraphics[width=\linewidth, frame]{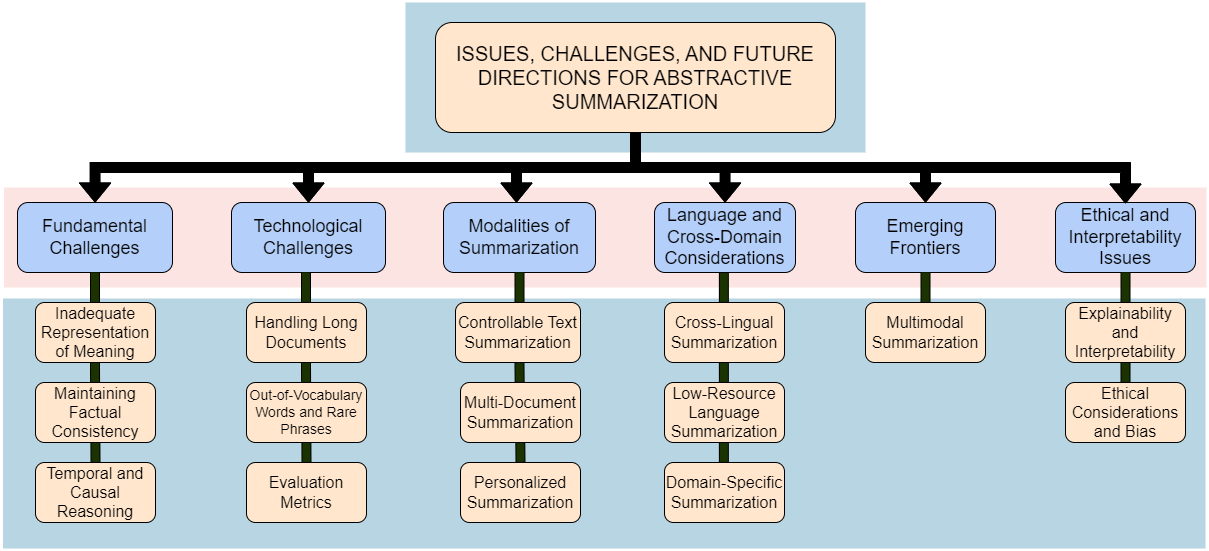}
    \caption{Taxonomy of Issues, Challenges, and Future Directions for Abstractive Summarization}
    \label{fig:TaxonomyIssue}
\end{figure*}

Abstractive text summarization has witnessed substantial progress, but challenges persist, and new research directions are emerging. This section provides a comprehensive overview of the current challenges, strategies to overcome them, and potential future research directions. See Figure \ref{fig:TaxonomyIssue} for the taxonomy associated with this section and its subclasses.

\subsection{Fundamental Challenges}
This section examines three fundamental challenges in abstractive summarization: inadequate representation of meaning, maintaining factual consistency, and temporal and causal reasoning as shown in Figure \ref{fig:TaxonomyIssue}. Regardless of advancements, models often struggle with capturing semantic nuances and the balance between completeness and conciseness. Integration of external knowledge structures and attention mechanisms show promise in improving meaning representation. Regarding factual consistency, condensation risks generating inaccurate or misleading summaries with severe implications, hence techniques leveraging knowledge bases, attribute control, and reinforcement learning are being examined to steer factual accuracy.

\subsubsection{Inadequate Representation of Meaning}


Despite advances, many summarization models struggle to adequately represent the meaning of the source text \cite{ji2023survey}. This inadequacy is not merely a result of model design but is deeply rooted in the inherent limitations of the underlying language models they are built upon. These foundational models, while powerful, sometimes struggle to grasp intricate semantic relationships, especially when the summarization task involves complex narratives or multifaceted arguments \cite{46111}. Among these foundational models, Large Language Models (LLMs), for instance GPT, have demonstrated remarkable capabilities in capturing semantic nuances \cite{digutsch2023overlap}. However, even these advanced models encounter challenges while trying to maintain the delicate balance between the original meaning of a text and the conciseness required for effective summarization. While LLMs are capable of processing large volumes of data and understand diverse linguistic patterns, they may be unable to accurately represent nuanced meanings or complex argumentative structures in the text \cite{kasneci2023chatgpt}. This challenge is 
further exacerbated by the complexity of the summarization task itself, which demands a delicate balance between conciseness and completeness.

To navigate these challenges, researchers have been exploring more advanced knowledge representation techniques. One promising approach is proposed by Banarescu et al. \cite{banarescu2013abstract}, which explores the use of leveraging external knowledge structures to aid in the summarization process. By integrating such advanced knowledge representations and the profound learning capabilities of LLMs, models can be better equipped to understand and reproduce the deeper semantic relationships present in the source text, leading to summaries that are not only concise but also rich in meaning and context.
Furthermore, attention mechanisms \cite{bahdanau2015neural} have been instrumental in helping models focus on relevant parts of the source text, thereby improving the representation of meaning. By weighing different parts of the input text based on their relevance, attention mechanisms allow models to capture more nuanced semantic relationships. Enhancing the capability of models to understand and incorporate complex semantic relationships and external knowledge structures will be crucial for the development of summarization models that produce summaries with greater depth, accuracy, and context awareness.

\subsubsection{Maintaining Factual Consistency}
Ensuring factual consistency in summaries is paramount \cite{cao2018faithful, tang2023aligning, lux2020truth}. As models condense vast amounts of information into concise summaries, there is a significant risk of distorting or misrepresenting the original content. Such distortions can generate summaries that, while grammatically correct, might convey misleading or factually incorrect information. The implications of these inconsistencies can be especially severe in areas like news dissemination, medical reporting, or legal documentation, where accuracy is crucial. 

Reinforcement learning (RL) techniques offer a potential solution to this challenge \cite{paulus2017deep}. In RL approaches, models are rewarded for generating factually consistent summaries. By establishing a reward mechanism that penalizes factual inconsistencies, models can be trained to prioritize factual accuracy over other aspects of summary generation.

As the field of abstractive summarization progresses, the focus on factual consistency will remain at the forefront, spurring innovations that emphasize truthfulness and accuracy in generated content. The development of more sophisticated RL techniques, which better incentivize factual consistency, is likely to be a key area of future research. Furthermore, integration of external knowledge sources, such as databases or knowledge graphs, and advanced reasoning capabilities will also play a crucial role in enhancing the factual accuracy of generated summaries.  By cross-referencing generated summaries with these trusted sources, models can validate the factual accuracy of their outputs. Another technique involves the use of attribute control in text generation \cite{logeswaran2018content}. Models can be steered to generate content that aligns with the ground truth by explicitly defining specific attributes or facts that the summary must adhere to. These advancements will contribute to the creation of models that provide factually accurate, coherent, and contextually appropriate summaries.

\subsubsection{Temporal and Causal Reasoning}
Incorporating temporal and causal reasoning in models is essential for coherent summaries \cite{zhao2022narrasum}. Recognizing this, there is a growing emphasis on integrating temporal and causal reasoning capabilities into modern summarization models. Techniques such as temporal logic offer a structured approach to capture and represent time-based relationships and sequences in texts \cite{allen1983maintaining}. By understanding the chronological order of events and their interdependencies, models can generate summaries that respect the natural progression of the narrative. On the other hand, causal models provide frameworks to discern cause-and-effect relationships within content \cite{pearl2000models}. However, there are inherent challenges in this domain. One of the primary issues is the ambiguity in natural language, where temporal and causal relationships might be implied rather than explicitly stated \cite{mirza2016catena}. This makes it challenging for models to consistently identify and represent these relationships. Additionally, the vastness and variability of real-world events, especially in domains like clinical narratives, mean that models often need to deal with incomplete or conflicting information, which can further complicate temporal and causal reasoning  \cite{bethard2016semeval}. 

A significant future direction involves the development of Hybrid Models. Combining rule-based approaches with deep learning can help better capture explicit and implicit temporal and causal cues \cite{tomer2020improving}. Another promising avenue is Knowledge Integration. Integrating external knowledge bases or ontologies that provide structured temporal and causal information can enhance the model's reasoning capabilities \cite{weikum2010information}. By identifying and representing these causal chains, summarization models can ensure that the underlying reasons and consequences of events are accurately reflected in the summaries. As the demand for high-quality, context-aware summaries grows, addressing these challenges and focusing on the aforementioned future directions will be pivotal in advancing the field and generating summaries that truly capture the essence and intricacies of the original content.

\subsection{Technological Challenges}
This section examines key technological challenges including handling long documents, OOV words, and rare phrases, and developing enhanced evaluation metrics as depicted in Figure \ref{fig:TaxonomyIssue}. Standard Seq2Seq models have trouble handling lengthy texts because of memory constraints and the challenge of capturing long-range dependencies. Tailored long-document models may prove vital, but hierarchical models and memory-augmented networks show potential in extracting the essence from large documents. Pre-trained models provide rich embeddings encompassing broad vocabularies to precisely process rare or uncommon terms. And new semantic similarity metrics are being explored to enhance surface-level ROUGE evaluation.

\subsubsection{Handling Long Documents}
Models, especially Seq2Seq architectures, find it challenging to summarize long documents \cite{liu2019text, gidiotis2020divide}. The inherent limitations of these models, such as memory constraints and the difficulty in capturing long-range dependencies, make them less adept at preserving the core essence of lengthy texts. Recognizing these challenges, researchers have been exploring alternative architectures and techniques. Hierarchical models, for instance, introduce a multi-level approach, where sentences are first encoded into sentence representations, which are then further encoded to generate the final summary \cite{nallapati-etal-2016-abstractive}. This layered approach aims to better manage the intricacies of lengthy texts. Similarly, memory-augmented neural networks enhance the model's capacity to remember and utilize information from earlier parts of the text, ensuring that crucial details are not lost in the summarization process \cite{miller-etal-2016-key}. 

While these methods have shown promise, there is a growing consensus in the research community about the need for models specifically tailored for long-document summarization. Such models would not just adapt existing techniques but would be fundamentally designed to handle the complexities and nuances of extensive texts \cite{grail2021globalizing}. As the digital world continues to produce vast amounts of lengthy content, from research papers to detailed reports, the demand for effective long-document summarization models will only intensify, making it a pivotal area for future research.

\subsubsection{Out-of-Vocabulary Words and Rare Phrases}
One of the persistent hurdles in natural language processing and machine learning is the effective handling of OOV words and rare phrases \cite{kouris2022abstractive, lochter2022multi}. These terms, which may not be present in the training vocabulary of a model, pose a significant challenge, especially when they carry critical information or context. The inability to process or generate such terms can lead to summaries that either omit essential details or resort to approximations, potentially compromising the accuracy and fidelity of the generated content. Managing OOV words and rare phrases involves various issues. Firstly, conventional word embedding techniques, for example, Word2Vec \cite{mikolov2013efficient} or GloVe \cite{pennington2014glove}, frequently neglect to give representations to words not seen during training. This limitation can bring about the loss of information when these words are encountered in real-world scenarios. Secondly, rare phrases, which comprise of various words, can be contextually rich, and their absence can cause a deficiency of nuanced meaning in the generated text \cite{luong2014addressing}. 

To explore this problem, the advances in powerful pre-trained models like BERT have given a pathway to address this issue. These models, trained on vast corpora, offer rich contextualized embeddings that capture a wide range of vocabulary, including many rare terms. By leveraging such pre-trained models, summarization systems can benefit from their extensive knowledge, ensuring that even less common words and phrases are handled with precision and context \cite{devlin2018bert}. As the demand for high-quality, comprehensive summaries grows, addressing the challenge of OOV words and rare phrases will remain central to the advancement of the field. The continued development and integration of advanced language models capable of understanding and representing a broader vocabulary are crucial for improving the accuracy and richness of generated summaries.

\subsubsection{Evaluation Metrics}

Evaluating the quality of generated summaries remains a challenge \cite{kryscinski2019evaluating, lloret2018challenging}. Traditional metrics, such as ROUGE, have been the cornerstone for evaluation due to their simplicity and ease of use. However, while ROUGE excels at doing surface-level comparisons between the generated summary and the reference, it often falls short of capturing deeper semantic similarities and nuances. This limitation becomes particularly evident when summaries, though semantically accurate, use phrasings or structures different from those in the reference text. Recognizing these shortcomings, the research community has been exploring alternative metrics. BERTScore \cite{zhang2019bertscore}, for instance, leverages the power of pre-trained LLMs to evaluate summaries based on contextual embeddings, offering a more nuanced measure of semantic similarity. Similarly, MoverScore \cite{zhao2019moverscore} introduced an approach by considering the movement of words in the generated summary, providing a different perspective on evaluating coherence and relevance. 

Although these newer metrics show considerable promise, the quest for the perfect evaluation metric is far from over. The dynamic and multifaceted nature of the summarization task demands continuous innovation in evaluation methodologies, urging researchers to delve deeper into the intricacies of summary quality and develop metrics that can holistically capture both form and essence \cite{ganesan2010opinosis}. Concurrently, human evaluation remains an invaluable tool for assessing the quality of summaries \cite{iskender2021reliability}. By incorporating human judgments, researchers can gain insights into aspects of summary quality that automated metrics might overlook, such as fluency, coherence, and overall informativeness. This multifaceted approach combining advanced, context-aware computational metrics with nuanced human evaluation will likely be the cornerstone of future research efforts in the quest to develop more accurate and meaningful summary evaluation methods.

\subsection{Modalities of Summarization}
Modalities of summarization are addressed in this section: controllable, multi-document, and personalized, as illustrated in Figure \ref{fig:TaxonomyIssue}. Controllable summarization allows customization to user requirements but maintaining coherence given rigid controls poses challenges. Multi-document synthesis requires aligning sources, eliminating redundancy, and representing perspectives in a balanced way - made difficult by large data volumes and conflicts. Personalized summarization aims to tailor summaries to individuals but handling dynamic user preferences is challenging. For all three domains, knowledge graphs and structured knowledge show the ability to enhance coherence, resolve conflicts, and anticipate user needs.

\subsubsection{Controllable Text Summarization}
There is a growing interest in controllable summarization, driven by the need for summaries tailored to diverse user needs and contexts \cite{fan2017controllable}. The CTRL model, a pioneering effort in controllable text generation, demonstrated the efficacy of control codes in directing content creation \cite{keskar2019ctrl}. However, it encountered challenges in achieving uniform performance across different controls and ran into potential overfitting issues. Controllable summarization meets this need by granting users the power to shape multiple facets of the summary, from its length and focus to its style and tone. This user-oriented methodology ensures summaries are not just succinct and logical but also pertinent to the context and tailored to individual preferences. One of the inherent challenges is to ensure that while a summary aligns with user-defined controls, it retains its coherence and fluency. See et al. \cite{see2019makes} delved into this delicate balance, revealing that rigid control parameters might sometimes yield summaries that, despite complying with the controls, compromise on coherence or overlook essential details. 

To realize this degree of personalization, future models must exhibit greater adaptability and greater sensitivity to data. Although controllable summarization aims to be user-centric, obtaining real-time feedback from users and incorporating it into the model can be challenging. This iterative feedback loop is crucial for refining and improving model outputs \cite{narayan2018don}. As the field progresses, focusing on the development of models that can better handle the balance between user controls and content coherence will be key. Enhancements in understanding user preferences and integrating real-time feedback will likely play a significant role in creating more sophisticated and user-responsive summarization systems.

\subsubsection{Multi-Document Summarization}

Although much research focuses on single-document summarization, multi-document summarization presents unique challenges \cite{narayan-etal-2018-ranking, lamsiyah2023unsupervised}. Unlike its single-document counterpart, multi-document summarization involves synthesizing information from multiple sources, often necessitating the alignment of documents, identification, and resolution of redundancies, contradictions, and varying perspectives. These complexities introduce unique challenges such as ensuring coherence in the face of diverse inputs and maintaining a balanced representation of all source documents. The enormous amount of information that needs to be processed during multi-document summarization is one of the main challenges. The amount of data increases rapidly with many documents, causing computational difficulties and extending processing times \cite{hong2014repository}. The possibility of conflicting information across documents presents another challenge. Finding the most precise or relevant information can be troublesome, particularly if the source text comprises different authors or viewpoints \cite{erkan2004lexrank}. Furthermore, the temporal part of the information can present difficulties. For instance, while summing up news articles, recent data may be more pertinent than older information, expecting models to have a sense of temporality \cite{wan2007manifold}. Large Language Models (LLMs) can play a crucial role. Their ability to process large volumes of text and understand complex linguistic patterns makes them well-suited for tackling the challenges of multi-document summarization \cite{ghadimi2023sgcsumm}. However, their application also introduces new dimensions to these challenges. For example, the computational resources required to process multiple lengthy documents using LLMs are significant, and the risk of perpetuating biases present in training data is heightened due to the models' extensive scope.

To address these issues, integrating knowledge graphs and structured knowledge representations has arisen as a promising strategy \cite{wang2022multi}. Knowledge graphs, with their interconnected nodes and connections, provide an organized system that can assist models, including LLMs, in understanding the connections between various text documents, recognizing key subjects, and generating summaries that capture the essence of the entire text document set \cite{chang2023survey}. Essentially, structured knowledge representations offer a deliberate method for coordinating and processing multi-document content, ensuring that the resulting summaries are comprehensive and well-structured. Recent advancements in attention mechanisms, particularly in transformer-based models, have also shown promise in effectively handling the intricacies of multi-document summarization \cite{liu2019hierarchical, perez2021multi}. As the demand for tools capable of distilling insights from vast and diverse data sets grows, multi-document summarization, strengthened by advanced knowledge structures and modern modeling techniques, including the use of LLMs, will undoubtedly play a pivotal role in shaping the future of information consumption.

\subsubsection{Personalized Summarization}
Traditional summarization techniques aim to distill the essence of texts for a general audience. However, with the increasing volume and diversity of information available, there is a growing recognition of the need for a more tailored approach: personalized summarization \cite{diaz2007user, vassiliou2023isummary}. This approach acknowledges that every reader comes with a unique background, preferences, and information needs. Instead of providing a one-size-fits-all summary, personalized summarization seeks to generate content summaries that resonate with individual users, emphasizing aspects most relevant to them. One primary issue is the dynamic nature of user preferences, which can evolve over time based on their experiences, interactions, and changing needs \cite{xiao2023chatgpt}. This dynamic nature requires models to be adaptive and responsive to ongoing user feedback. Incorporating user-specific knowledge, such as their reading history, preferences, or feedback, can provide valuable insights into what they value in a summary \cite{narayan2018don}. Furthermore, leveraging knowledge graphs offers another layer of personalization \cite{ji2021survey}. These graphs can map out intricate relationships between different pieces of information, allowing the generation of summaries that not only cater to a user’s current interests but also anticipate their future queries or areas of interest.

LLMs like GPT and BERT greatly improve personalized summarization due to their advanced natural language understanding \cite{richardson2023integrating}. They can adapt effectively to user preferences by analyzing interactions and feedback, enabling them to generate summaries that are increasingly tailored to individual user profiles over time \cite{van2023clinical}. Moreover, a significant future direction in this realm is Adaptive Learning. Building models that can learn and adapt from continuous user feedback, especially those based on LLMs, will be crucial in ensuring that the summaries remain aligned with evolving user preferences \cite{zhang2008adasum, peng2023check}. As the digital landscape becomes increasingly user-centric, addressing these challenges and focusing on adaptive learning, augmented by the capabilities of LLMs, will be instrumental in ensuring that readers receive concise, relevant, and engaging summaries tailored just for them. These models will need to not only understand and reflect individual user preferences but also continually adapt to changing user needs and preferences over time. The integration of advanced techniques such as knowledge graphs to provide a deeper, more context-aware level of personalization will also be a key area of future research.

\subsection{Language and Cross-Domain Considerations}
As shown in Figure \ref{fig:TaxonomyIssue}, this section examines cross-lingual, low-resource language, and domain-specific challenges. Cross-lingual translation risks meaning loss and linguistic structure differences hence advanced neural translation shows potential. For low-resource languages, limited datasets hinder supervised learning while unique nuances may be neglected. Data scarcity is addressed by techniques like transfer learning, data augmentation, and multilingual models. Domain adaptation poses challenges due to specialized lexicons and data limitations, hence domain knowledge graphs and transfer learning show potential.

\subsubsection{Cross-Lingual Summarization}
Cross-lingual summarization, which involves generating concise summaries in a target language different from the source, is becoming increasingly vital due to the globalization of information \cite{wang2022survey}. This task faces challenges such as potential loss of meaning during translation and the intricacies of different linguistic structures \cite{wang2022survey}. The quality of the source document is important because ambiguities or poor structure can make the summarization process more difficult \cite{wan2010towards}. One common solution to this problem is to summarize the machine-translated content after the source documents have been translated into the target language \cite{wan2010cross}. However, this approach occasionally introduces mistakes because errors in the translation stage can trickle down to the summarization stage. A more direct approach is provided by advanced neural machine translation models combined with summarization methods, which guarantee that the essence of the original content is retained in the summarized output \cite{cao2017improving}. To train and refine cross-lingual summarization models, parallel corpora—datasets that combine text in one language with its translation in another—have become increasingly popular. By utilizing such corpora, models are able to learn complex language mappings and generate summaries that are both precise and fluent \cite{tiedemann2009news}. Another promising path is the creation of multilingual models, like those trained using the MASK-PREDICT method \cite{ghazvininejad2019mask}. These models, trained on data from multiple languages, possess the capability to understand and generate text across a wide linguistic spectrum. 

In an era of increasing globalization and a rising need for accessible information in multiple languages, the progress made in cross-lingual summarization, supported by machine translation methods, will have a crucial impact on overcoming language barriers and promoting global comprehension. The development of advanced neural translation models that can effectively combine with summarization methods to retain the essence of the original content will be a key area of future research. Additionally, the utilization of parallel corpora and the creation of robust multilingual models will play a significant role in advancing the field. This will ensure that the generated summaries are not only linguistically accurate but also maintain the semantic integrity and fluency of the original content across different languages.

\subsubsection{Low-Resource Language Summarization}

In our linguistically diverse world, major languages like English, Chinese, and Spanish dominate the digital realm, leaving many low-resource languages underrepresented in computational models \cite{ruder2019survey}. This imbalance challenges natural language processing tasks, such as text summarization. The limited availability of parallel corpora hinders supervised training \cite{zoph2016transfer}, and nuances unique to these languages can be overlooked, resulting in potentially inaccurate summaries \cite{prediger2019one}. Pre-training models on related languages, like using Spanish to aid Catalan summarization, can be effective \cite{le2019flaubert}. Data augmentation, including back-translation, improves training data \cite{sennrich2015improving}, while transfer learning bridges the data gap by adapting models from high-resource to low-resource languages \cite{zoph2016transfer}. Unsupervised methods, like Google's ``zero-shot translation," use high-resource languages as intermediaries \cite{johnson2017google}. 

The future of summarization for low-resource languages will likely involve cross-lingual pre-trained models, such as multilingual BERT \cite{pires2019multilingual}, and few-shot learning techniques that maximize limited data \cite{wang2023adapting}. Collaborations among communities, linguists, and technologists can produce more annotated datasets \cite{bird2003seven}, and models that respect cultural nuances will be essential \cite{bender2011achieving}. As the digital age strives for inclusivity, these strategies will ensure that every language and culture is represented. The continuous development of models trained on a broader range of languages and the incorporation of cultural and linguistic nuances will be crucial in making the benefits of advanced summarization technologies accessible to all language communities.

\subsubsection{Domain-Specific Summarization}

In the vast landscape of information, content varies not only in style but also in substance, depending on the domain from which it originates. Whether it is the intricate jargon of medical literature, the precise terminology of legal documents, or the nuanced language of academic research, each domain has its unique lexicon and conventions \cite{yogan2016review}. Generic summarization models can struggle with these domain-specific nuances. A significant challenge is the scarcity of specialized training data, with limited datasets in areas like bioinformatics compared to general news \cite{jones2007automatic}. Additionally, the dynamic nature of many fields means that domain knowledge can quickly become outdated, requiring models to be continually updated with the latest information \cite{radev1997generating}. To address this, there is a shift towards domain-specific knowledge graphs that capture field-specific relationships and terminologies \cite{abu2021domain}. By incorporating these into summarization models, summaries become more domain-aware and resonate with experts. 

A significant future direction involves the application of Transfer Learning. Leveraging models pre-trained on general domains and fine-tuning them on domain-specific datasets can help bridge the data gap in specialized fields \cite{howard2018universal}. As the demand for specialized content summarization rises, these strategies will ensure summaries remain accurate and contextually apt. The continual adaptation of summarization models to incorporate up-to-date domain knowledge and terminologies will be crucial in maintaining the relevance and accuracy of the summaries. The development of advanced domain-specific knowledge graphs and the enhancement of model adaptability to dynamic domain changes will play a pivotal role in refining the capability of summarization technologies to cater to specific field requirements, ensuring that the generated summaries are both informative and expertly aligned with domain-specific needs.

\subsection{Emerging Frontiers}
This section examines the emerging frontier of multimodal summarization as depicted in Figure \ref{fig:TaxonomyIssue}, which aims to summarize both textual and visual data to provide comprehensive representations. Standard techniques often fall short with such multifaceted inputs. The challenges encompass aligning textual and visual modalities effectively and ensuring coherence. Promising directions involve techniques designed to integrate cross-modal data through unified textual-visual algorithms.

\subsubsection{Multimodal Summarization}
With the rise of multimedia content, summarizing visual and textual data is becoming increasingly crucial \cite{jangra2021survey}. Traditional text-based summarization techniques, while effective for pure textual content, may fall short when faced with the task of distilling information from both visual and textual sources. This is due to the inherent complexity of visual data, which often contains rich, non-linear information that does not always have a direct textual counterpart \cite{li2017scene}. Challenges arise in aligning visual and textual modalities, handling the vast variability in visual content, and ensuring that the generated summaries maintain coherence across both modalities \cite{zhang2018retrospective}.

Recognizing this gap, there is an emerging focus on multimodal summarization, which seeks to generate concise representations that capture the essence of both visual and textual elements.  Techniques that integrate visual and textual content are at the forefront of this endeavor \cite{li2022fusing}. By developing algorithms that can understand and interrelate images, diagrams, videos, and text, it becomes possible to generate summaries that truly reflect the composite nature of multimedia content. As the digital landscape continues to evolve, with visuals playing an ever-increasing role, addressing these challenges and focusing on the aforementioned future directions will be paramount in ensuring that users receive a holistic understanding of the content they consume. Advancements in algorithms capable of effectively integrating and summarizing cross-modal data through unified textual-visual representations will be crucial in meeting the growing need for comprehensive, coherent multimodal summaries.

\subsection{Ethical and Interpretability Issues}
This section examines ethics and interpretability in abstractive summarization as shown in Figure \ref{fig:TaxonomyIssue}. Since models run the risk of sustaining data biases, tools for mitigating bias and transparent knowledge graphs are promising. Interpretability is also paramount for trust in high-stakes applications. By mapping model reasoning, knowledge graphs facilitate interpretability, and integrated methods that further enhance transparency.

\subsubsection{Explainability and Interpretability}

The black-box nature of many advanced models, including Large Language Models (LLMs), has raised significant concerns, especially when these models are used in high-stakes domains such as healthcare, finance, or legal decision-making \cite{lipton2018mythos, thirunavukarasu2023large}. In these contexts, a model's outputs must be not only accurate but also understandable to stakeholders. The need for understanding how and why specific decisions are made is crucial for establishing trust, ensuring accountability, and, in some cases, meeting regulatory compliance requirements. The interpretability of LLMs, with their complex and often opaque internal mechanisms, presents unique challenges in abstractive summarization \cite{luo2023reasoning}. Discerning how these models arrive at specific summarization decisions can be difficult, particularly when the summaries need to be explainable and justifiable, as in legal or medical contexts. To address these challenges, recent research has focused on integrating explainability features such as ``attention highlights" into these models. These highlights, which closely mirror the model’s decision-making process and align with the user’s mental model of the task, have been shown to significantly increase user trust and efficiency \cite{norkute2021towards}. However, ensuring the explainability and interpretability of models, including LLMs, becomes a paramount concern. While simpler models are often more interpretable, they may not achieve the performance levels of more complex models, presenting a trade-off between complexity and interpretability \cite{ribeiro2016should, luo2023reasoning}. Many current interpretability methods provide post-hoc explanations, which might not accurately reflect the actual decision-making process of the model \cite{doshi2017towards}. Researchers are exploring more transparent knowledge representation methods and utilizing knowledge graphs to demystify the inner workings of these models and provide tangible and visual representations of their decision-making processes \cite{rajabi2022knowledge}. Furthermore, the development of methods like Layer-wise Relevance Propagation (LRP) and other deep learning visualization techniques is underway to provide clearer insights into the intricate patterns used by LLMs in generating summaries \cite{bassi2024improving}.

A significant future direction is Integrated Interpretability, where future models, including LLMs, might be designed with interpretability as an intrinsic feature of their architecture, rather than relying solely on post-hoc methods \cite{chen2018learning}. As we continue to integrate AI-driven solutions into critical sectors of society, addressing these challenges and focusing on integrated interpretability will be instrumental in ensuring these technologies, particularly LLMs, are adopted responsibly and ethically. The development of models with in-built interpretability will not only enhance trust and transparency but also ensure that these models are aligned with the needs and values of society, contributing to the responsible and ethical use of AI in decision-making processes.

\subsubsection{Ethical Considerations and Bias}

The generation of unbiased and objective summaries is of paramount importance. As models often learn from large amounts of data, they can inadvertently adopt and perpetuate the biases present within these data sets \cite{hovy2016social}. Such biases, whether they are related to gender, race, culture, or other social factors, can skew the content of summaries, leading to misrepresentations and reinforcing harmful stereotypes. Addressing this challenge requires a multifaceted approach. The use of transparent knowledge representation methods can help in bias mitigation \cite{kamishima2012fairness}, by making the decision-making processes of models more interpretable, allowing researchers and users to identify, understand, and rectify potential biases in the generated summaries. 

A significant future direction in this realm is the development of Bias Detection and Correction Tools. The creation of tools that can automatically detect and correct biases in summaries will be crucial. These tools can provide real-time feedback to models, allowing them to adjust their outputs accordingly \cite{sun2019mitigating}. As society becomes increasingly aware of the ethical implications of artificial intelligence, ensuring fairness and eliminating biases in text summarization will be crucial in building trustworthy and equitable systems. The advancement of these tools, along with the continuous improvement of transparent knowledge representation methods, will play a pivotal role in addressing ethical considerations and reducing bias in AI-driven text summarization technologies, contributing to the creation of more just and fair AI systems.

In conclusion, although considerable progress has been made, challenges persist in abstractive text summarization, and the field offers many opportunities for innovation. By addressing these challenges and exploring the highlighted research directions, we can pave the way for more effective and reliable summarization systems.

\section{Concluding Remarks}
In conclusion, this survey paper has provided a comprehensive and well-structured analysis of the field of abstractive text summarization. We discussed state-of-the-art methods, highlighting the variety of approaches and the rapid advances being made. Furthermore, we have identified the areas that require improvement for more efficient summarization models by identifying and analyzing the current issues and challenges in the field. Our examination of strategies for overcoming these challenges has highlighted the significance of integrating knowledge and other techniques in the creation of abstractive summarization models.

Additionally, we have identified areas of interest that have the potential to significantly advance the field and have explored promising avenues of future research. Enhancing factual consistency, developing cross-lingual and multilingual summarization systems, concentrating on domain-specific summarization, dealing with noisy data, and enhancing long-document summarization are a few of these research directions.

We have also provided vital comparison tables across techniques in each summarization category, offering insights into model complexity, scalability and appropriate applications.

We hope that by providing this well-organized overview of abstractive text summarization, researchers and practitioners will be motivated to tackle the challenges and pursue the future research directions presented in this paper. It is essential to keep focusing on the development of more efficient, trustworthy, and beneficial summarization models that can be used for a variety of purposes as the field grows.

\section*{Acknowledgement}
During the preparation of this work, the author(s) used ChatGPT in order to improve flow of writing in approximately 10\% of the document. After using this tool/service, the authors reviewed and edited the content as needed and take full responsibility for the content of the publication.



\bibliographystyle{elsarticle-num} 
\bibliography{main.bib}





\end{document}